%% file: acl_latex.tex
\pdfoutput=1

\documentclass[11pt]{article}

\usepackage[preprint]{acl}

\usepackage{times}
\usepackage{latexsym}

\usepackage[T1]{fontenc}

\usepackage[utf8]{inputenc}

\usepackage{microtype}

\usepackage{inconsolata}

\usepackage{graphicx}
\usepackage{comment}
\usepackage[utf8]{inputenc} 
\usepackage[T1]{fontenc}    
\usepackage{hyperref}       
\usepackage{url}            
\usepackage{booktabs}       
\usepackage{amsfonts}       
\usepackage{nicefrac}       
\usepackage{microtype}      
\usepackage{xcolor}         
\usepackage{booktabs}
\usepackage{dcolumn}
\usepackage{array}
\usepackage{graphicx}
\usepackage{amsmath}
\usepackage{multirow}
\usepackage{natbib}
\usepackage{enumitem}
\usepackage[most]{tcolorbox}
\usepackage{xcolor}
\usepackage{pifont}
\usepackage{listings}
\usepackage{subfigure}
\usepackage{caption}
\usepackage{todonotes}
\definecolor{nb}{HTML}{006EB8}
\usepackage{hyperref}
\hypersetup{
  colorlinks   = true,    
  urlcolor     = nb,    
  linkcolor    = nb,    
  citecolor    = nb      
}

\definecolor{codegreen}{rgb}{0,0.6,0}
\definecolor{codegray}{rgb}{0.5,0.5,0.5}
\definecolor{codepurple}{rgb}{0.58,0,0.82}
\definecolor{backcolour}{rgb}{0.95,0.95,0.92}
\definecolor{codeblue}{rgb}{0.25,0.5,0.75}
\definecolor{darkblue}{rgb}{0.1,0.1,0.6}
\definecolor{lightgrey}{rgb}{0.9,0.9,0.9}

\lstdefinestyle{mystyle}{
    language=Python,
    backgroundcolor=\color{lightgrey},   
    commentstyle=\color{codegreen},
    keywordstyle=\color{codeblue}\bfseries,
    numberstyle=\tiny\color{codegray},
    stringstyle=\color{codepurple},
    breakatwhitespace=false, 
    basicstyle=\ttfamily\tiny,
    breaklines=true,                 
    captionpos=b,                    
    keepspaces=true,                 
    numbers=left,                    
    numbersep=5pt,                  
    showspaces=false,                
    showstringspaces=false,
    showtabs=false,                  
    tabsize=2,
    classoffset=1, 
    morekeywords={bfs, Solution, findShortestCycle, TestCase, Output}, 
    keywordstyle=\color{darkblue}\bfseries,
    classoffset=0
}

\newcommand{\methodname}{{\tt{SSD}}}
\newcommand{\datasetname}{{\tt{GraphEval36K}}}

\title{\datasetname{}: Benchmarking Coding and Reasoning Capabilities of Large Language Models on Graph Datasets}

\definecolor{deepblue}{rgb}{0.0, 0.0, 0.55}

\author{
  Qiming Wu\thanks{These authors contributed equally.}, Zichen Chen\footnotemark[1], Will Corcoran, Misha Sra, Ambuj Singh \\
  University of California, Santa Barbara \\
  \texttt{\{qimingwu,zichen\_chen,wcorcoran,sra,ambuj\}@cs.ucsb.edu}
}

\begin{document}
\maketitle
\begin{abstract}
Large language models (LLMs) have achieved remarkable success in natural language processing (NLP), demonstrating significant capabilities in processing and understanding text data. However, recent studies have identified limitations in LLMs' ability to manipulate, program, and reason about structured data, especially graphs. We introduce \datasetname{}\footnote{\datasetname{} is available at \url{https://grapheval36k.github.io/}, under MIT license.}, the \textbf{first} comprehensive graph dataset, comprising 40 graph coding problems and 36,900 test cases to evaluate the ability of LLMs on graph problem-solving.
Our dataset is categorized into eight primary and four sub-categories to ensure a thorough evaluation across different types of graphs. We benchmark ten LLMs, finding that private models outperform open-source ones, though the gap is narrowing. We also analyze the performance of LLMs across directed vs undirected graphs, different kinds of graph concepts, and network models.
Furthermore, to improve the usability of our evaluation framework, we propose Structured Symbolic Decomposition (\methodname), an instruction-based method designed to enhance LLM performance on complex graph tasks. Results show that \methodname{} improves the average passing rate of GPT-4, GPT-4o, Gemini-Pro and Claude-3-Sonnet by 8.38\%, 6.78\%, 29.28\% and 25.28\%, respectively.
\end{abstract}

\input{intro}

\begin{figure*}[htb]
    \centering
    \includegraphics[width=0.9\linewidth]{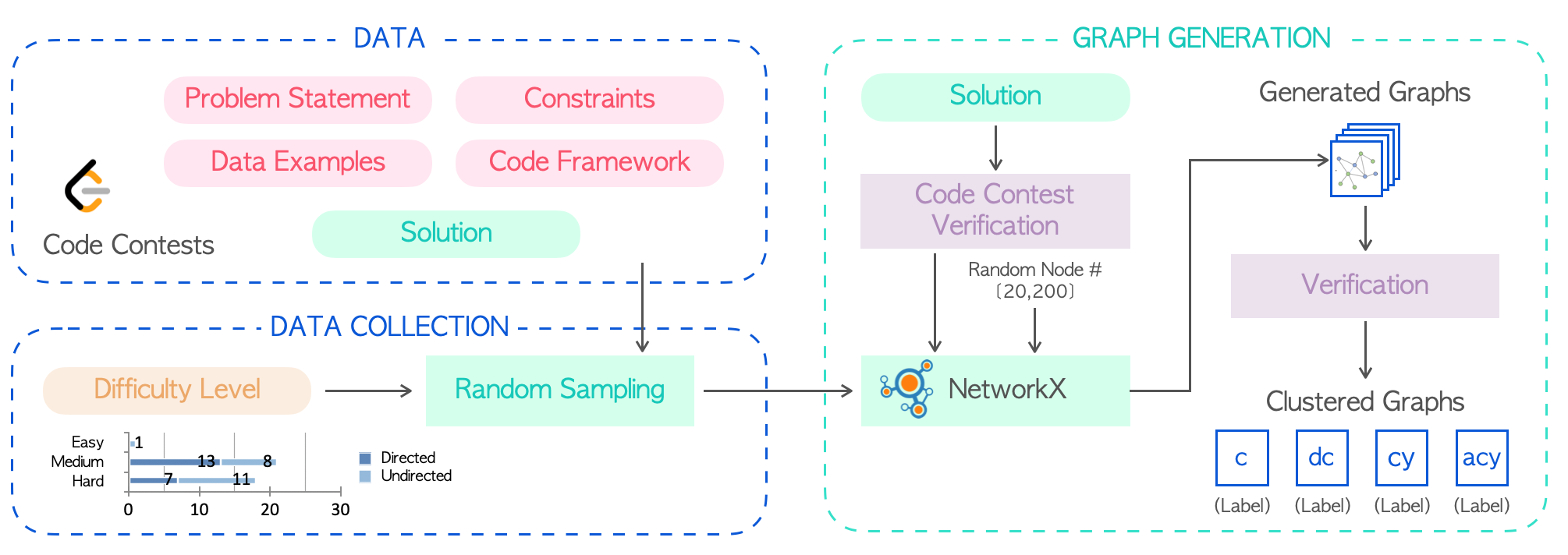}
    \caption{
    The \datasetname{} dataset is constructed through a pipeline that begins with data collection from code contests (LeetCode). Next, problems are randomly sampled according to their difficulty levels, and corresponding graphs are generated using NetworkX. These graphs are then clustered and labeled based on whether they are connected (c), disconnected (dc), cyclic (cy), or acyclic (acy). Verification steps ensure labeling accuracy, though the exact labels may vary depending on each graph's characteristics.}\label{fig:data_collection}
\end{figure*}

\section{Related Work}

\subsection{LLMs on Graph Problem-Solving}

Recent research on using LLMs for graph-related tasks follows two approaches: (1) natural language interaction, and (2) code generation.

In the natural language interaction approach, LLMs are provided with a graph and they generate answers based on their understanding of the graph structure. Current research finds that they perform adequately on basic graph problems, however, their performance declines on more complex graph problems~\citep{jin2023large, wang2024can, liu2023evaluating, guo2023gpt4graph}. Inspired by the Chain-of-Thought (CoT) method~\citep{wei2022chain,kojima2022large}, researchers propose a step-by-step reasoning framework to help LLMs solve graph problems in a more structured way~\citep{chai2023graphllm, wang2024can, liu2023evaluating, guo2023gpt4graph, zhang2023llm4dyg, fatemi2023talk, sun2023think}. Notably, when LLMs are provided with additional example solutions, their performance improves. However, the improvement is marginal~\citep{zhang2023llm4dyg, fatemi2023talk, chai2023graphllm}.

In the code generation approach, recent research demonstrates that LLMs exhibit significantly improve problems-solving abilities through code generation, instead of relying solely on natural language responses~\citep{Suzgun2022ChallengingBT, Liang2023HolisticEO, Hendy2023HowGA}. Researchers transform complex problems into code problems to facilitate effective interaction and enhance the performance of LLMs~\citep{madaan-etal-2022-language, zhang-etal-2023-causal, bi2024codekgc, dong2022corrpus,yan2023codescope, cai2024codegraph}.

\subsection{LLMs vs. Traditional Graph Machine Learning}

Traditional graph machine learning (ML) models, such as Graph Neural Networks (GNNs)~\citep{kipf2016semi, velickovic2017graph}, demonstrate strong performance in tasks like node classification~\citep{xiao2022graph}, link prediction~\citep{zhang2018link}, and graph classification~\citep{zhang2018end}. However, they are designed for specific tasks with certain architectures~\citep{garg2020generalization}. In contrast, LLMs have the potential to generalize across different graph-related tasks, without requiring specific tuning~\citep{jin2024large,sun2023graph}. In this work, we aim to evaluate the ability of LLMs to solve graph problems by generating correct code.

\subsection{Comparison with Existing Datasets}

\input{table2}

Table~\ref{tab:dataset_comparison} presents a comparative comparison of \datasetname{} with existing datasets designed to evaluate LLMs in graph-related tasks. Unlike previous datasets, which focus primarily on specific aspects of graph reasoning, such as algorithmic problem solving (CLRS-30~\citep{velivckovic2022clrs}), logical inference (ProofWriter~\citep{tafjord2021proofwriter}, PrOntoQA~\citep{misra2023prontoqa}) or hybrid graph analysis (HGB~\citep{li2023hybrid}), \datasetname{} offers a \textbf{comprehensive evaluation framework} that includes both graph coding and reasoning. Furthermore, while datasets such as BIG-Bench~\citep{srivastava2022beyond} and GraCoRe~\citep{yuan2024gracore} include graph-related reasoning tasks, they are often constrained to small-scale graphs or heterogeneous structures, limiting their applicability to broader graph problem-solving.

\section{Dataset Construction}

\begin{figure}
    \centering
    \includegraphics[width=0.9\linewidth]{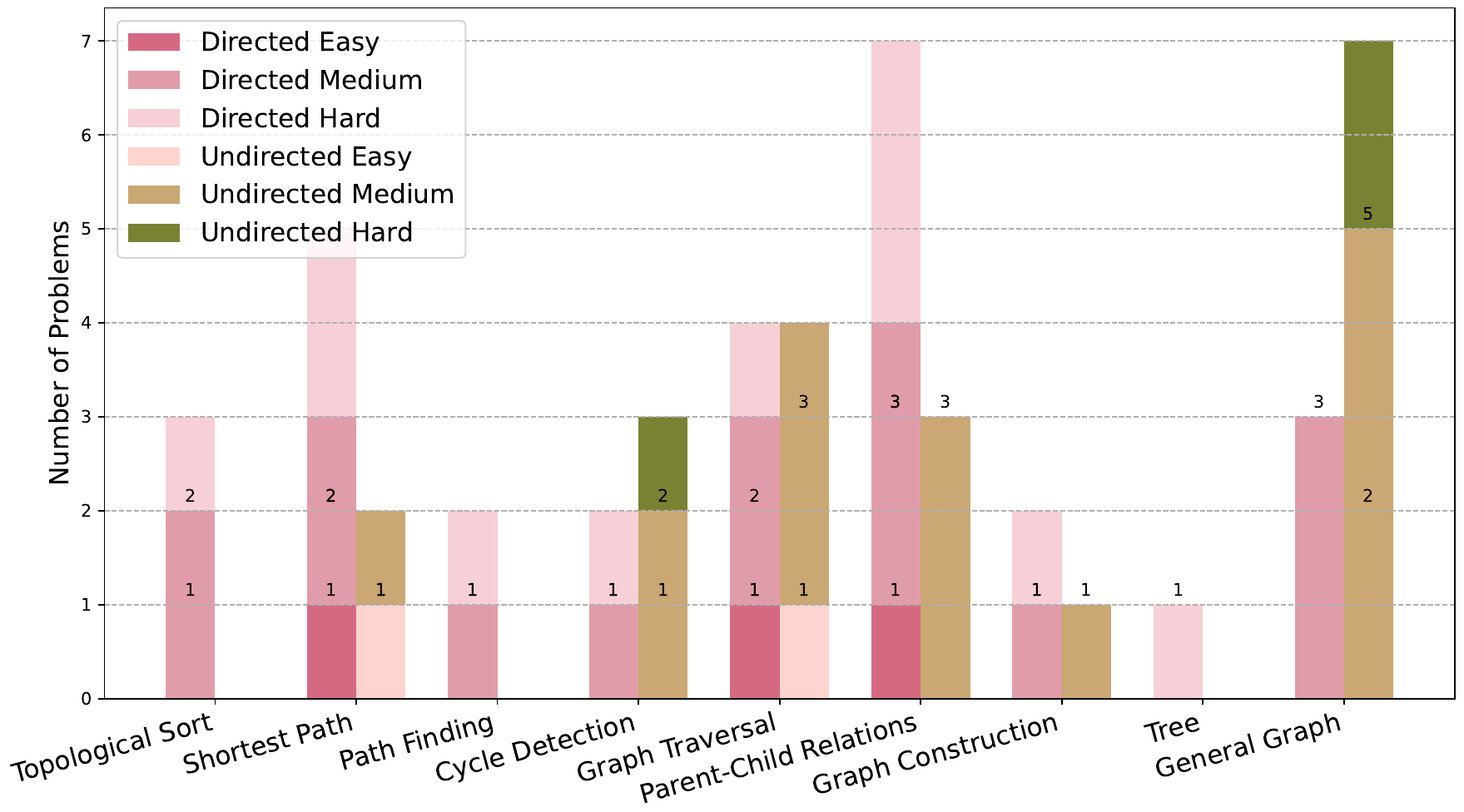}
    \caption{Distribution of graph problems on concepts and difficulty levels.}
    \label{fig:distribution_of_leetcode}
\end{figure}

An overview of the dataset construction pipeline is illustrated in Figure \ref{fig:data_collection}. The two major steps consist of (a) selection of graph problems with varying degrees of user-defined difficulty levels and graph concepts, (b) generation of graphs of different types.

\subsection{Problem Collection}
We collect a total of 40 graph data structure problems from the LeetCode\footnote{\url{https://leetcode.com/tag/graph/}}, comprising 20 undirected and 20 directed graph problems. Most of these problems are recently released, which minimizes the likelihood of their inclusion in the training sets of the verification LLMs. The distribution of the problems across difficulty levels and graph concepts is presented in Figure \ref{fig:distribution_of_leetcode}. For each problem, we collect: problem statement, input/output examples, data constraints, and code framework.

\subsection{Graph Generation}
Consider a graph $G = (V, E)$, where $V$ denotes the set of vertices and $E$ denotes the set of edges in the graph. We classify graphs into eight main categories: Sparse Graph, Planar Graph, Regular Graph, Dense Graph, Complete Graph~\citep{diestel2024graph}, Small-world Graph~\citep{watts1998collective}, Erdos-Renyi Graph~\citep{erdos1960evolution} and Power-law Graph~\citep{barabasi1999emergence}. Below, we provide definitions of these graphs and how they are generated. 

\paragraph{Sparse Graph}  A graph $G = (V, E)$ is considered sparse if the number of edges $|E|$ is much less than the maximum possible number of edges~\citep{bollobas2011sparse}: 
\begin{equation}
    \left\{ \begin{array}{lll}
    |E| \ll \frac{|V|(|V|-1)}{2} \quad \text{(undirected graph),} \\
    \\ |E| \ll |V|(|V|-1) \quad \text{(directed graph)}.
\end{array}\right.
\end{equation}

\paragraph{Planar Graph} A planar graph can be drawn without edge intersections, except at vertices~\citep{schnyder1989planar}. For a finite, connected planar graph with $F$ faces following Euler's formula, we have:
\begin{equation}
    |V| - |E| + |F| = 2.
\end{equation}

\paragraph{Regular Graph} A graph $G$ is $k$-regular if every vertex has the same degree $k$~\citep{stanic2017regular}.

\paragraph{Dense Graph} For an undirected graph, the density \( D \) is given by $
D_{\text{undirected}} = \frac{2|E|}{|V|(|V| - 1)}$~\citep{lee2008pebble}. For a directed graph, the density \( D \) is $D_{\text{directed}} = \frac{|E|}{|V|(|V| - 1)}$. In our experiment, we fix \( D = 0.7 \).

\paragraph{Complete Graph} A graph $G$ is complete if there is an edge between every pair of distinct vertices~\citep{pirnot2001mathematics}.

\paragraph{Small-World Graphs} Small-world graphs, generated via the Watts-Strogatz model~\citep{watts1998collective}, use parameters $n$ (nodes), $k$ (nearest neighbors), and $p$ (rewiring probability), where $k$ is randomly chosen between $2$ and $\frac{n}{2}$, and $p$ between $0.1$ and $0.3$. Starting with a ring lattice where each node $i$ connects to its $k$ nearest neighbors, each edge $(u, v)$ is rewired with probability $p$ to a random node. The resulting graph has a logarithmic path length $L(G)$ and a high clustering coefficient $C(G)$.

\paragraph{Erdos-Renyi Graphs} Erdos-Renyi graphs~\citep{erdos1960evolution} follow the $G(n, p)$ model, where $n$ is the number of nodes, and $p$ is the probability of forming an edge between any two nodes. Each edge $(u, v)$, where $u \neq v$, is included in $E$ with probability $p$.

\paragraph{Power-Law Graphs} Power-law graphs are generated using the Barabasi-Albert (BA) model with parameters $n$ (nodes) and $m$ (edges per new node)~\citep{barabasi1999emergence}, where $m$ is randomly chosen between 1 and 10. New nodes connect to $m$ existing nodes via preferential attachment.
This process yields a scale-free network with a Power-law degree distribution:
\[
P(k) \sim k^{-\gamma}.\vspace{-5pt}
\]

\begin{figure}[tb]
\vspace{-5pt}
    \centering
    \includegraphics[width=0.9\linewidth]{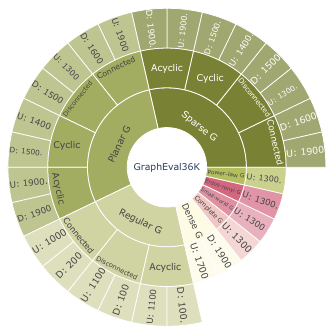}
    \caption{Structure of \datasetname. ``U'' denotes undirected graphs, ``D'' denotes directed graphs, with numbers indicating the count of cases in each category. The graphs are classified into eight main categories: sparse, planar, regular, dense, complete, Small-world, Erdos-Renyi, and Power-law. Some are further divided into four sub-categories: connected, disconnected, cyclic, and acyclic. Sub-categories may vary based on the characteristics of the main categories. Detailed dataset analysis is shown in Appendix~\ref{section: analysis_of_dataset}.}

    \label{fig:graph-classification}
\end{figure}

\paragraph{Graph Generation} We use the ``NetworkX''~\citep{networkx} to generate graph samples. By leveraging it, we construct both directed and undirected graphs according to our classification results (as shown in Figure \ref{fig:graph-classification}). We generate 100 graph samples for each sub-category, where the number of vertices in each graph is randomly chosen between 20 and 200. This enables us to create the dataset with varying levels of complexity.

\input{table1.tex}

\subsection{LLMs under Consideration} 
We evaluate ten LLMs in this study: Claude-3-Sonnet~\citep{claude3}, Gemini-Pro~\cite{reid2024gemini}, GPT-3.5~\citep{gpt3}, GPT-4, GPT-4o~\citep{achiam2023gpt}, Llama-3-8b, Llama-3-70b~\citep{dubey2024llama}, Mixtral-8x7b~\citep{jiang2024mixtral}, Qwen2.5-Coder-32B~\citep{hui2024qwen2} and DeepSeek-V3~\citep{liu2024deepseek}. The first five models are private, while the latter five are open-source. These models are selected for their diversity in scale, architecture, and relevance within the research community. Our objective is to assess their graph-solving capabilities and compare their performance on the \datasetname{} dataset.

\begin{figure}[t!]
    \centering
    \subfigure[Sparse Graphs]{
        \includegraphics[width=0.22\textwidth]{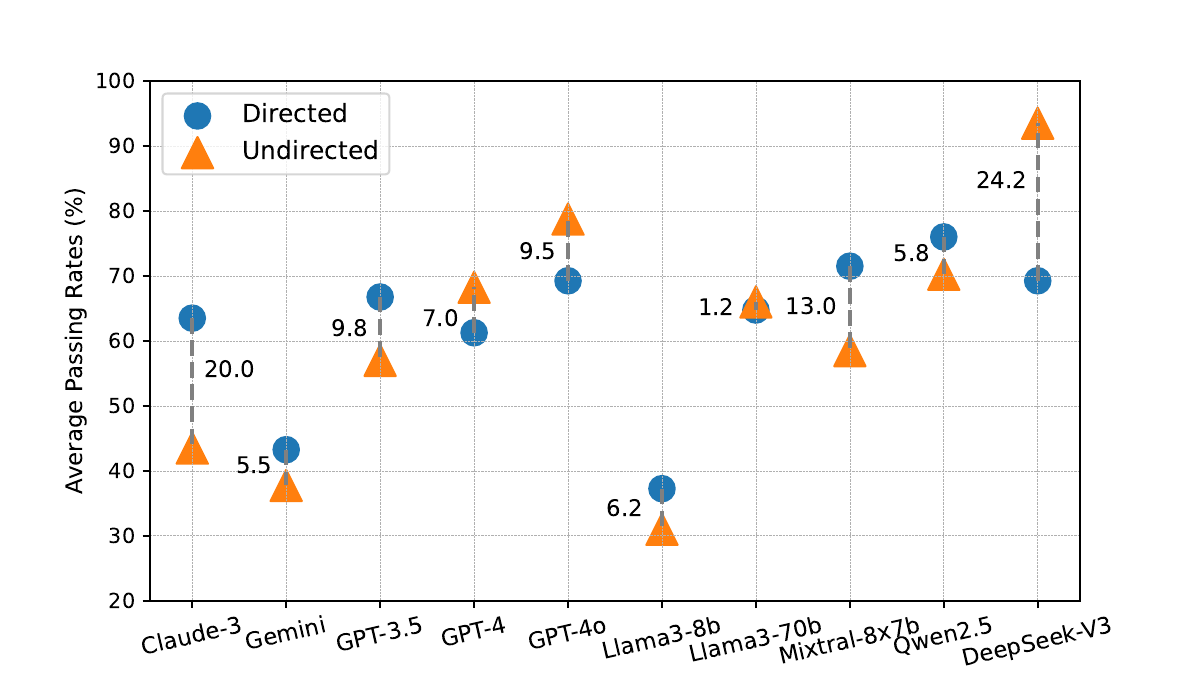}\label{fig:sparse_dataset_eval}
    }
    \subfigure[Planar Graphs]{
        \includegraphics[width=0.22\textwidth]{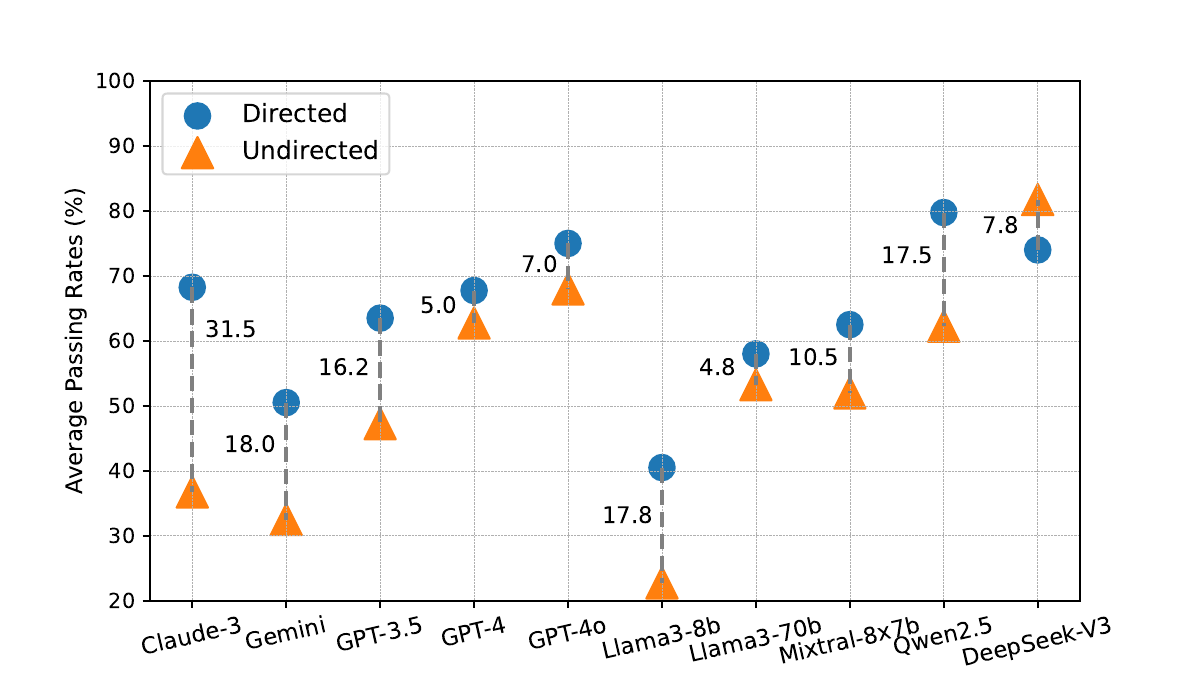}\label{fig:planar_dataset_eval}
    }
    \caption{Average passing rate on sparse and planar graphs. We mark the absolute value of the difference between results of directed and undirected graphs.}\label{fig:res_sparse_planar}
\end{figure}

\section{Evaluation of LLMs on Graph Problems}
\begin{figure}[t]
    \centering
    \subfigure[Easy Concepts]{
        \includegraphics[width=0.42\columnwidth]{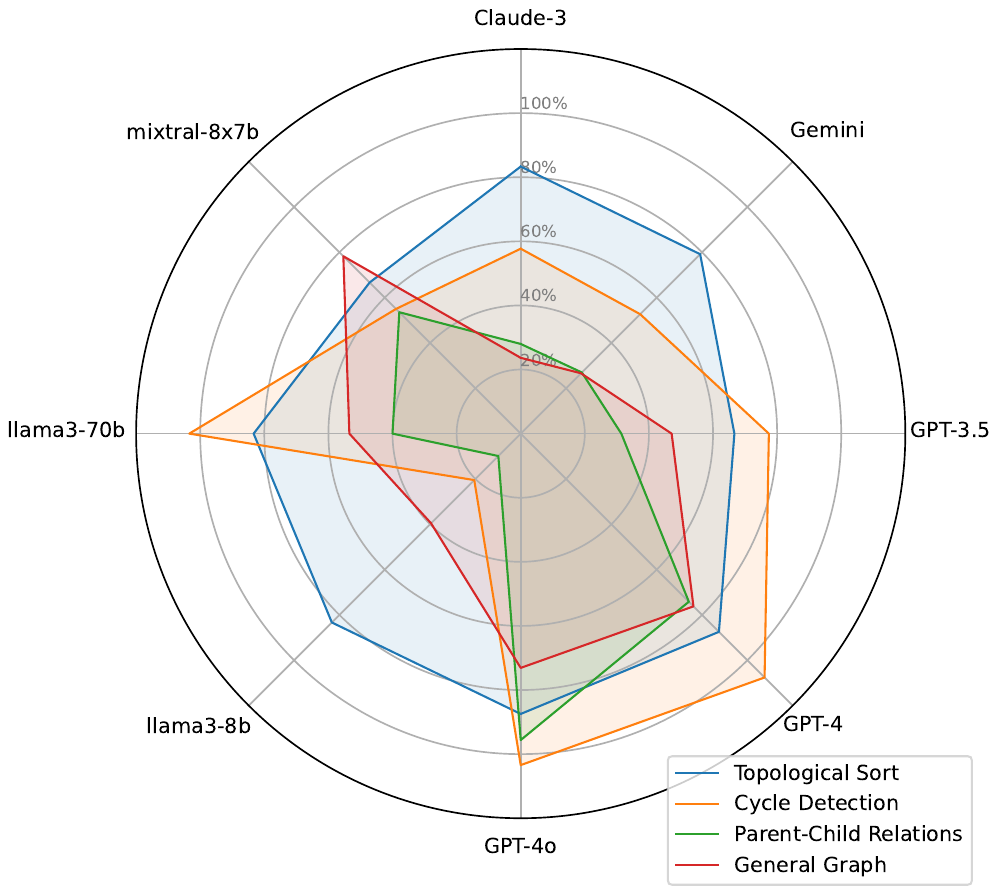}\label{fig:part-2}
    }
    \subfigure[Difficult Concepts]{
        \includegraphics[width=0.42\columnwidth]{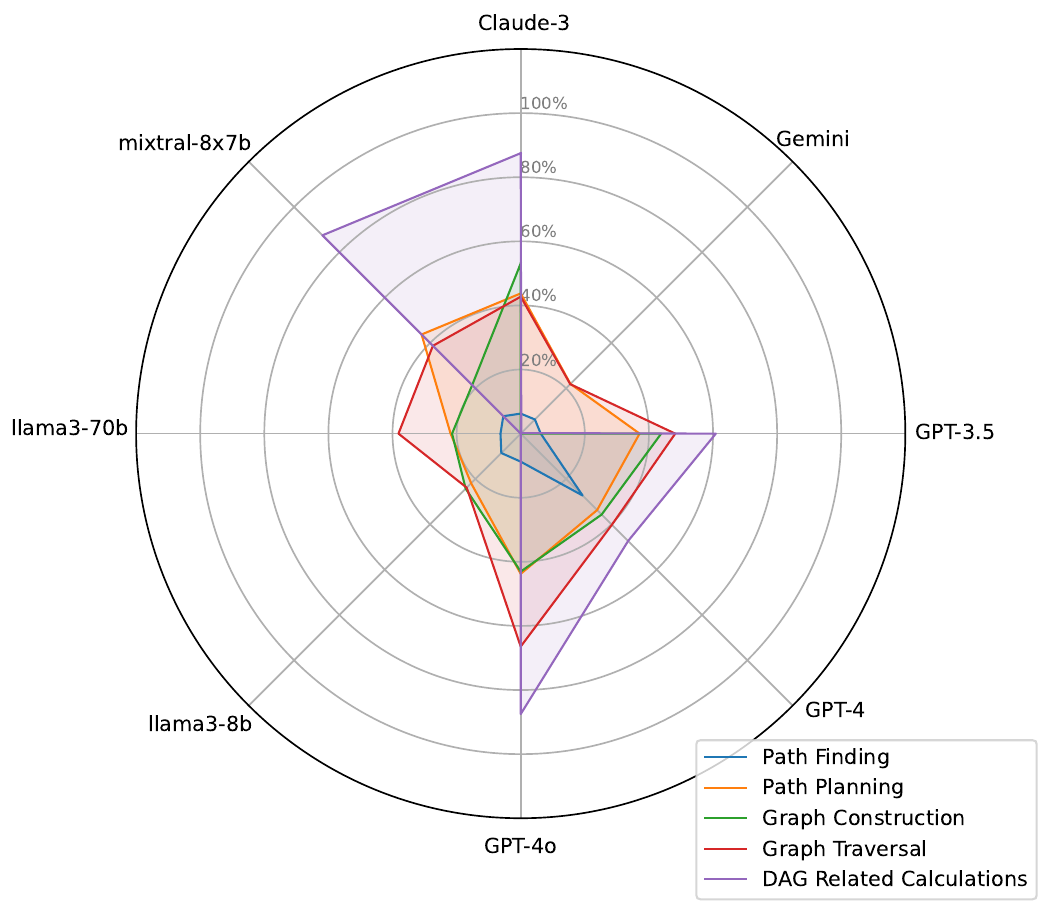}\label{fig:part-1}
    }
    \caption{Passing rate of LLMs on graph concepts. Part (a) presents the passing rate on easy concepts, and part (b) presents the passing rate on difficult concepts.}
    \label{fig:part-1 and part-2}
\end{figure}

\begin{figure}[t!]
    \centering
    \includegraphics[width=0.95\linewidth]{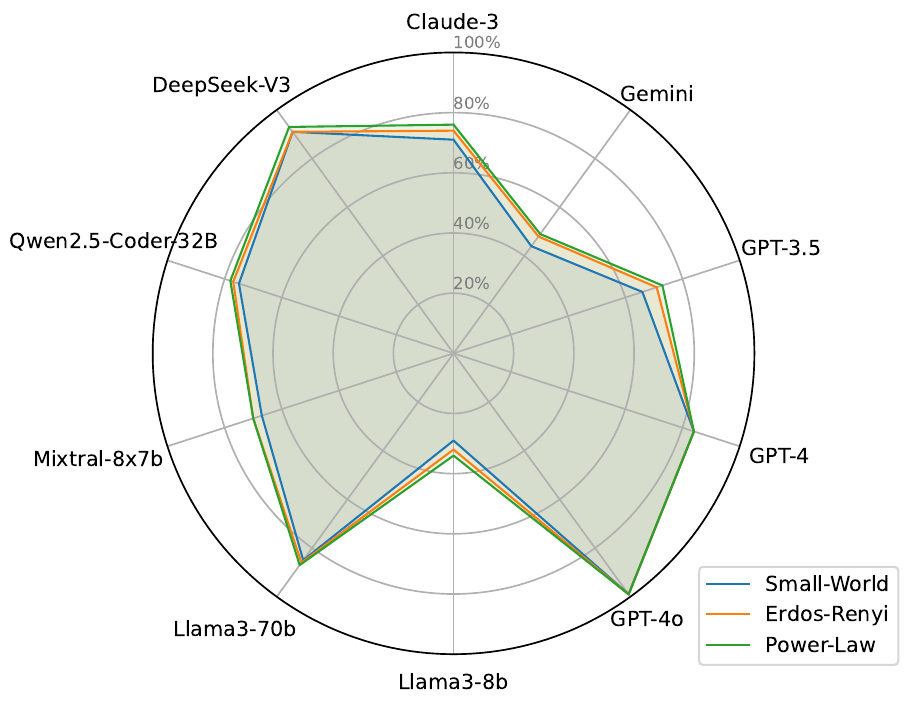}
    \caption{Evaluation results of LLM-generated code on complex graphs: Small-world, Erdos–Renyi, and Power-law.}
    \label{fig:3-new-graph-eval-results}
\end{figure}

In this section, we evaluate the LLMs' capability in
solving graph problems.
We first discuss the broad results of evaluating LLMs on graph problems (Table \ref{tab:test_dataset}), and then perform three analytical studies. First, we evaluate how well LLMs understand core graph theory concepts in their problem-solving strategies (Figure~\ref{fig:part-1 and part-2}). Second, we analyze the impact of directed vs. undirected graphs on LLM coding performance (Table \ref{tab:test_dataset} and Figure~\ref{fig:res_sparse_planar}). Finally, we evaluate model generalization to complex graphs, including Small-world, Erdos-Renyi, and Power-law graphs (Figure~\ref{fig:3-new-graph-eval-results}).

In Table~\ref{tab:test_dataset}, we observe that private models consistently outperform open-source models in all graph categories, especially GPT-4 and GPT-4o. The performance gap is notable in complex graph types such as dense and complete graphs, where private models achieve passing rates close to 100\%, while open-source models like Mixtral-8x7b perform in a lower range of 70-80\%. However, Qwen2.5-Coder-32B and DeepSeek-V3 have narrowed the performance gap. They perform better than private models on sparse and planar graphs.

\paragraph{Results on Graph Problem Concepts} 
We classify graph problems into nine concepts:
Topological Sort, Cycle Detection, Parent-Child Relations, General Graph, Path Finding, Path Planning, Graph Construction, Graph Traversal and DAG-Related Calculations. As shown in Figure \ref{fig:part-1 and part-2}, 
LLMs perform better on the concepts in Figure \ref{fig:part-1}, and face more difficulty with those in Figure \ref{fig:part-2}.
For example, LLMs consistently show strong performance on Topological Sort, with Claude-3-sonnet, GPT-4, and GPT-4o achieving rates above 80\%. LLMs generally excel at structured problems involving clear hierarchical relationships. However, Claude-3-sonnet and Gemini-pro perform poorly on Path Finding, with passing rates of 6.25\% and 21.94\%, respectively. In contrast, GPT-4 and GPT-4o show a stronger result on DAG-related Calculations, with GPT-4o achieving 87.50\%. The results highlight the strengths and weaknesses of LLMs across different graph concepts, emphasizing the need for further research to improve performance on more complex graph problems.

\paragraph{Results on Directed and Undirected Graphs}
Most LLMs perform better on directed graphs, but it is hard to draw firm conclusions due to potential differences in problem complexity between directed and undirected graphs. As shown in Table~\ref{tab:test_dataset} and Figure \ref{fig:res_sparse_planar}, the performance gap between private and open-source models is more notable on directed graphs, indicating strong capabilities of private LLMs in handling directed graph complexities. Furthermore, private models show a larger performance gap between directed and undirected graphs compared to open-sourced models. However, an interesting observation is that DeepSeek-V3 demonstrates superior performance on undirected graphs compared to directed graph cases. Notably, DeepSeek-V3 outperforms all evaluated LLMs on undirected sparse and planar graph cases, highlighting its strength in these specific graph structures. Further analyses can be found in Appendix~\ref{sec:appendix-usage} and~\ref{sec:appendix-llm-d&und}.

\paragraph{Results on Complex Graphs} The evaluation results of LLMs across three practical graph types, Small-world, Erdos–Renyi, and Power-law, are summarized in Figure \ref{fig:3-new-graph-eval-results}. GPT-4o demonstrates the highest performance across all graph types, maintaining a passing rate of 99\%. 
DeepSeek-V3 follows closely with a passing rate of 93\%. Additionally, GPT-4, Qwen2.5-Coder-32B, and Llama-3-70B exhibit strong performance, with passing rates ranging from 84\% to 87\%, demonstrating their robust capability in solving graph-related problems. In contrast, Gemini-pro and Llama-3-8b show the lowest passing rates across all graph types, particularly for Small-world graphs where it achieves only 29\%. Claude-3-sonnet, GPT-3.5, and Mixtral-8x7b occupy a middle ground, showing moderately high but varied performance across different graph types, with rates ranging from 66\% to 76\%. 
The results indicate that while models like GPT-4o and GPT-4 excel at generating code to solve practical graph problems, performance is variable across different LLMs. This highlights the need for further fine-tuning or architectural improvements to enhance the code generation capabilities of weaker models, particularly in complex graph scenarios.

\begin{table*}[htbp]
    \centering
    \renewcommand{\arraystretch}{0.6} 
    \begin{tabular}{@{}lcc|cc|cc|cc@{}}
    \toprule
    \multirow{2}{*}{} & \multicolumn{2}{c|}{\textbf{GPT-4}} & \multicolumn{2}{c|}{\textbf{GPT-4o}} & \multicolumn{2}{c|}{\textbf{Gemini-Pro}} & \multicolumn{2}{c}{\textbf{Claude-3-Sonnet}} \\ \cmidrule(l){2-9} 
                                      & \textbf{Vanilla}  & \textbf{+SSD}  & \textbf{Vanilla}   & \textbf{+SSD}  & \textbf{Vanilla}   & \textbf{+SSD}  & \textbf{Vanilla}   & \textbf{+SSD}   \\ \midrule
    Path Finding                      & 80.25              & 88.83          & 82.15               & 88.76          & 0                   & 0.37           & 85.24               & 95.56           \\
    Path Planning                     & 100.00                & 100.00            & 100.00                 & 100.00            & 36.78               & 66.67          & 68.96               & 100.00             \\
    Graph Construction                & 16.13              & 19.29          & 22.32               & 25.49          & 12.97               & 19.04          & 31.44               & 36.75           \\
    DAG Related                       & 85.50              & 100.00            & 90.35               & 100.00            & 0                   & 2.44           & 54.88               & 84.15           \\ \midrule
    Directed                          & 51.25              & 64.92          & 91.67               & 92.31          & 15.54               & 40.71          & 66.67               & 75.00           \\
    Undirected                        & 89.87              & 92.00          & 84.35               & 92.00          & 25.61               & 64.00          & 52.91               & 83.20           \\ \midrule
    Small-World                       & 100.00                & 100.00            & 100.00                 & 100.00            & 0                   & 68.50          & 50.00               & 98.50           \\
    Erdos-Renyi                       & 98.00              & 100.00            & 100.00                 & 100.00            & 33.33               & 79.33          & 64.67               & 95.67           \\
    Power-Law                         & 100.00                & 100.00            & 100.00                 & 100.00            & 33.33               & 80.00          & 66.67               & 100.00             \\ \bottomrule
    \end{tabular}
    
    \caption{Passing rates of four LLMs (GPT-4, GPT-4o, Gemini-Pro, and Claude-3-Sonnet) across different graph problems and types of graph samples. We randomly sampled six problems, a total of 6,000 graph samples from our dataset, to compare model performance with and without (vanilla) the proposed \methodname{} method.}
    \label{tab:ssd_on_concepts_and_3_new_graphs}
\end{table*}

\input{method}

\section{Conclusion}
In this work, we introduce \datasetname{}, the first graph dataset designed to benchmark LLMs' graph reasoning abilities through coding challenges. It includes 40 graph problems and 36,900 graph test cases, covering a range of difficulty levels and graph concepts. To enhance LLM performance, we propose \methodname{}, an instruction-based approach aimed at improving reasoning capabilities on graph-related problems. Our experiments demonstrate the effectiveness of \datasetname{} and \methodname{} in analyzing LLM performance across different graph types and concepts. Our findings highlight potential areas for improvement and identify the graph coding tasks where LLMs perform well and where they face challenges.



\input{limitation}

\bibliography{ref}

\clearpage
\newpage

\appendix

\section{Analysis of \datasetname}\label{section: analysis_of_dataset}
In this section, we analyze our dataset from several perspectives: data distributions, the complexity of graph test cases, and the time and memory usage during evaluations on \datasetname.

\begin{figure}[htbp]
    \centering
    \includegraphics[width=\linewidth]{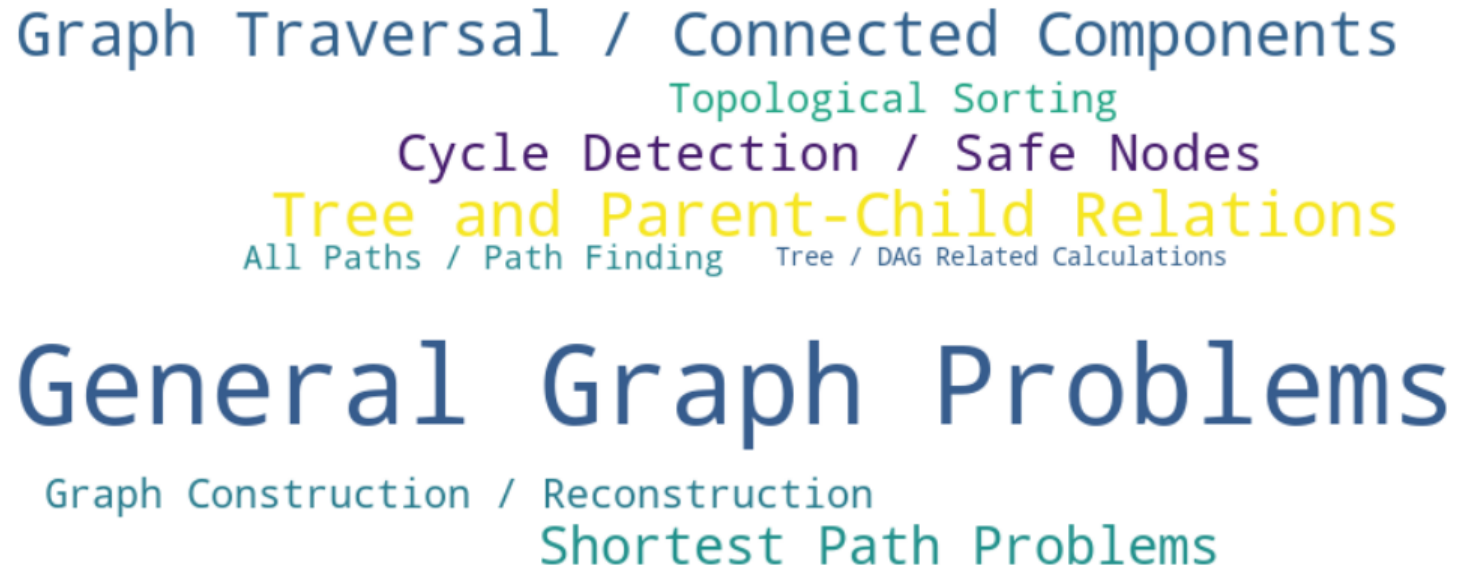}
    \caption{Word cloud of the distribution of data structure problems on concepts.}
    \label{fig:word-cloud-leetcode}
\end{figure}

\subsection{Data Distributions}
\begin{figure}[htbp]
    \centering
    \includegraphics[width=0.9\linewidth]{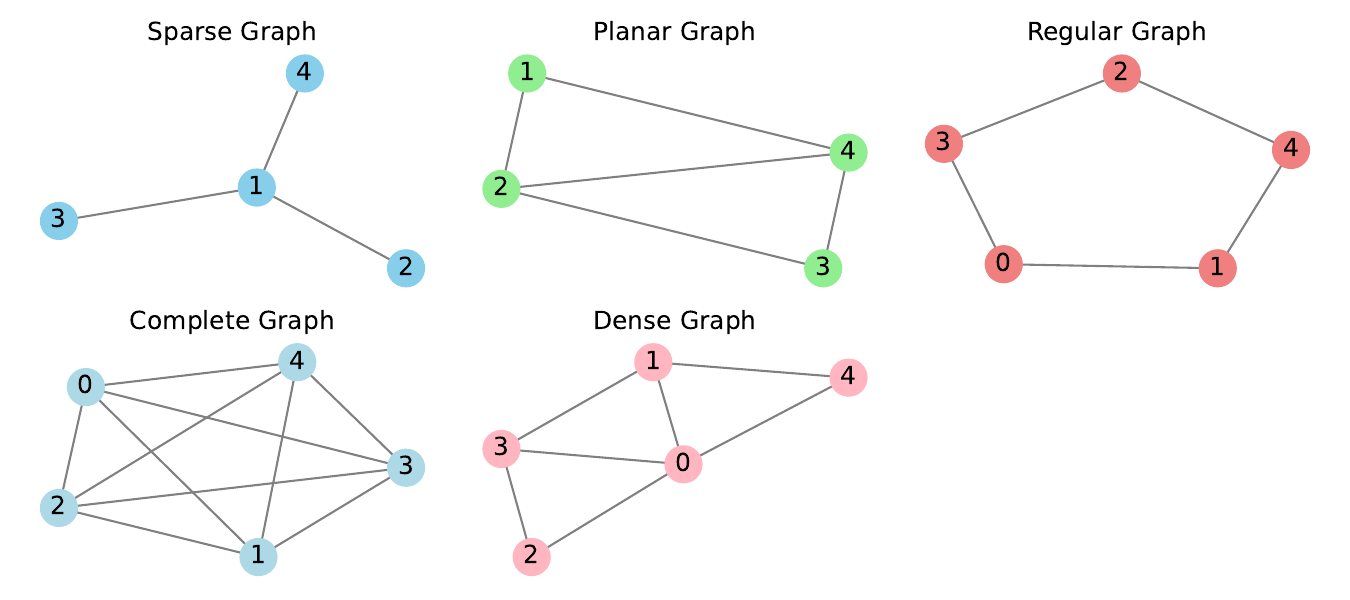}
    \caption{Graph Samples. We use NetworkX (\url{https://networkx.org/}) to plot the graph samples of eight main categories in our dataset.}
    \label{fig:graph_samples}
\end{figure}

\begin{figure}[htbp]
    \centering
    \includegraphics[width=\linewidth]{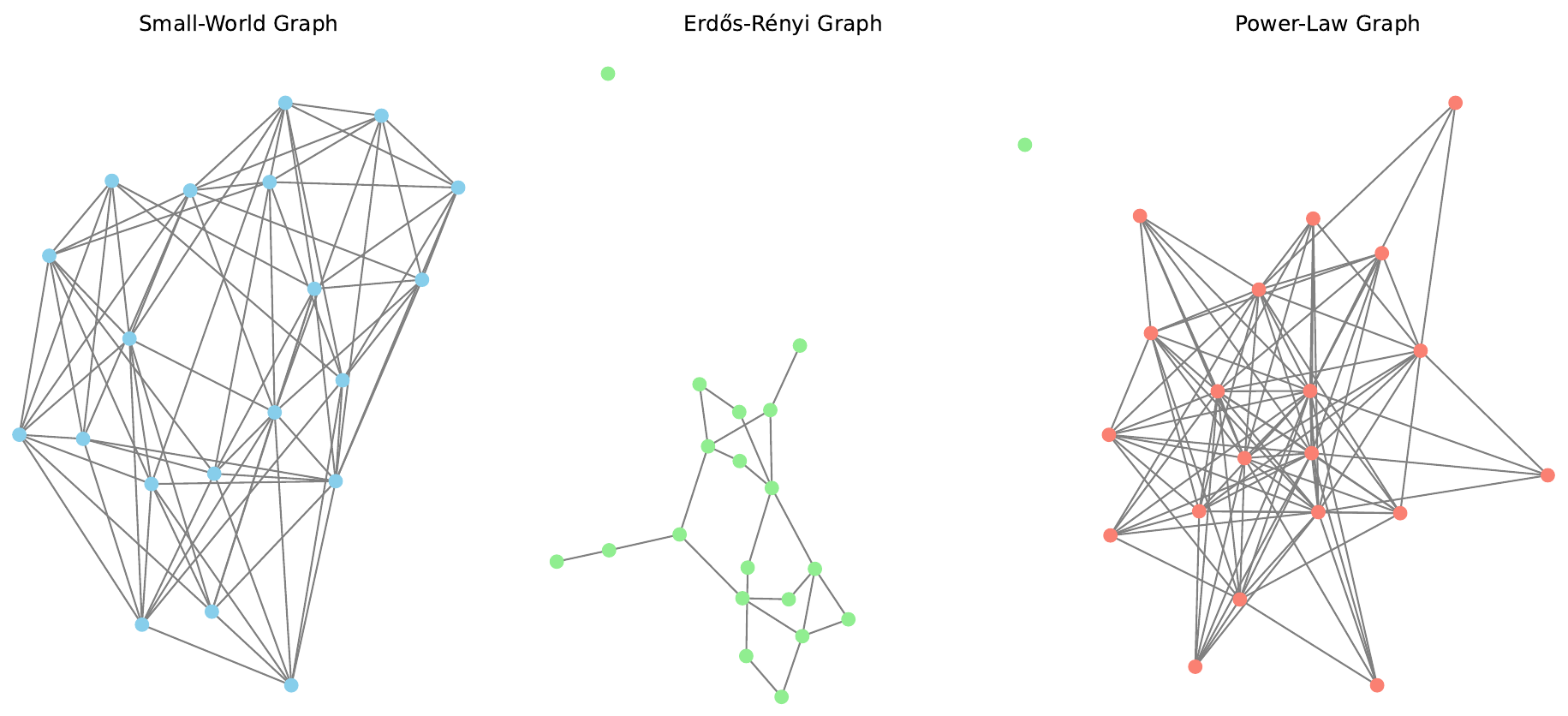}
    \caption{Graph Samples of Small-world, Erdos-Renyi and Power-law graphs. We use NetWorkX to plot these graph samples with the number of nodes to be 20.}
    \label{fig:3-new-graph-samples}
\end{figure}

We begin by enumerating the graph test cases. There are 13,400 samples in the directed graph category and 23,500 in the undirected graph category. Therefore, \datasetname has 36,900 graph samples in total. Additionally, we examine the distribution of problems across different difficulty levels and concepts, as illustrated in Figure \ref{fig:distribution_of_leetcode}. Our dataset, \datasetname, includes eight main graph categories (as shown in Figure \ref{fig:graph_samples} and Figure \ref{fig:3-new-graph-samples}, sparse, planar, regular, dense, complete, Small-world, erdos-renyi and Power-law) and four subcategories (connected, disconnected, cyclic, and acyclic) within each main category. For each subcategory, we generated ten graph test cases. Due to the specific characteristics of the main categories, our final dataset comprises \textbf{13,400} graph samples for directed graphs and \textbf{23,500} for undirected graphs. These samples are designed to evaluate and improve the graph reasoning abilities of LLMs. Additionally, we generate the word cloud (Figure \ref{fig:word-cloud-leetcode}) of our dataset by different graph problem concepts.

\begin{figure*}[ht!]
    \centering\includegraphics[width=\textwidth]{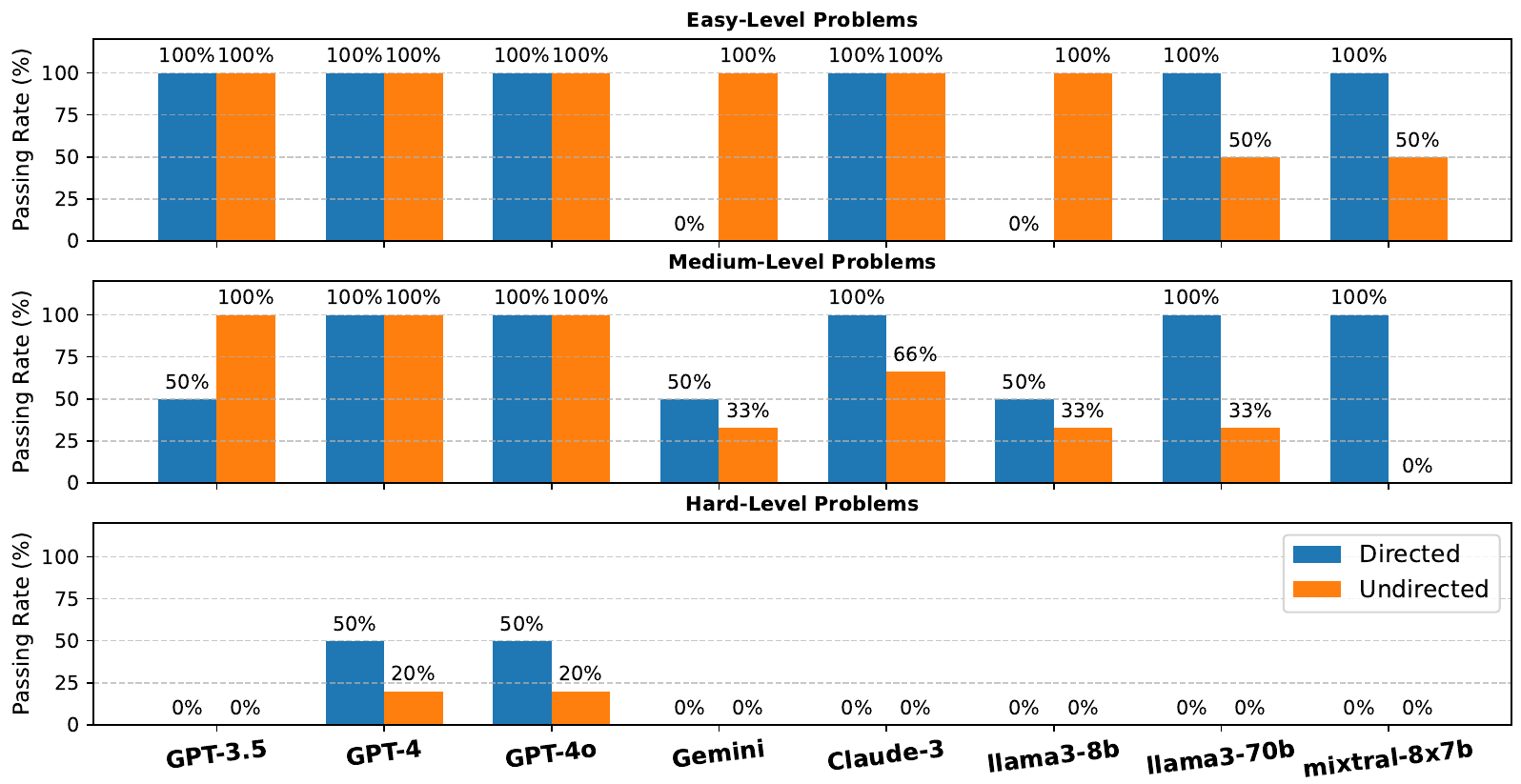}
    \caption{Evaluation results of LLMs on the LeetCode platform. This figure shows the passing rates of LLMs on selected graph data structure problems, categorized into 3 easy-level problems, 5 medium-level problems, and 9 hard-level problems.}\label{fig:online_leetcode_result}
\end{figure*}

The graph problems were collected from LeetCode (\url{https://leetcode.com/tag/graph/}), and we calculated the distribution of problems across different difficulty levels\footnote{The difficulty level is defined by LeetCode.}. Given the near-perfect performance of LLMs on easy-level data structure problems, we focused primarily on medium- and hard-level problems. 

\begin{table*}[htbp]
\centering
\renewcommand{\arraystretch}{0.8}
\begin{tabular}{@{}lcccc@{}}
\toprule
Graph Type          & \textbf{\begin{tabular}[c]{@{}c@{}}Average Number \\ of Nodes\end{tabular}} & \textbf{\begin{tabular}[c]{@{}c@{}}Average Number \\ of Edges\end{tabular}} & \textbf{\begin{tabular}[c]{@{}c@{}}Average \\ Degree\end{tabular}} & \textbf{\begin{tabular}[c]{@{}c@{}}Average \\ Degree Variance\end{tabular}} \\ \midrule
Directed Sparse     & 103.97                                                                       & 268.21                                                                      & 4.79                                                               & 3.72                                                                        \\
Directed Planar     & 41.09                                                                       & 43.64                                                                       & 2.28                                                               & 1.18                                                                        \\ \midrule
Undirected Sparse   & 88.51                                                                       & 89.16                                                                       & 1.95                                                               & 0.72                                                                        \\
Undirected Planar   & 70.52                                                                       & 71.89                                                                       & 1.98                                                               & 0.74                                                                        \\
Undirected Regular  & 82.96                                                                       & 201.16                                                                      & 5.01                                                               & 0.64                                                                        \\
Undirected Complete & 73.04                                                                       & 4019.89                                                                     & 72.05                                                              & 0.0 
\\ \midrule
Small-world & 101.22 & 1660.99 & 26.05 & 4.33
\\
Erods-Renyi & 103.43 & 824.91 & 12.74 & 10.65
\\
Power-law & 103.17 & 543.54 & 10.18 & 46.09
\\ \bottomrule
\end{tabular}
\caption{Complexity Analysis of Graphs. We include the average number of nodes and edges, average degree, and average degree variance to analyze the graph test cases in \datasetname.}
\label{tab:complexity_analysis}
\end{table*}

\subsection{Dataset Statistics} 
To analyze the complexity of the graph test cases in \datasetname, we include several key metrics: the average number of nodes and edges, the average degree, and the average degree variance. These metrics provide a comprehensive understanding of the structural characteristics of the graphs in our dataset, allowing for a more detailed assessment of the complexity and diversity of the test cases. The results are summarized in Table \ref{tab:complexity_analysis}.

To begin with, we first introduce the evaluation metrics. Consider $N_i$ to be number of nodes in the $i$-th graph, $E_i$ represents the number of edges in the $i$-th graph. Then, average number of nodes $\overline{N}$ across $m$ graph test cases is 
\begin{equation}
    \overline{N} = \frac{1}{m} \sum_{i=1}^{m} N_i,
\end{equation}
and average number of edges $\overline{E}$ across $m$ graph test cases is
\begin{equation}
    \overline{E} = \frac{1}{m} \sum_{i=1}^{m} E_i.
\end{equation}

The degree of a node in a graph is the number of edges connected to it. The average degree of a graph is the mean degree of all its nodes. For an undirected graph with $N$ nodes and $E$ edges, the average degree $\overline{d}$ is
\begin{equation}
    \overline{d} = \frac{2E}{N}.
\end{equation}
For a directed graph, the average degree is
\begin{equation}
    \overline{d} = \frac{E}{N}.
\end{equation}

Across multiple graphs, the average degree is the mean of the average degrees of each graph. The degree variance measures the variability of the degrees of the nodes in a graph. For a graph with $N$ nodes and degrees $d_1, d_2, \ldots, d_N$, the variance of the degrees is calculated as:
\begin{equation}
    \text{Var}(d) = \frac{1}{N} \sum_{i=1}^{N} (d_i - \overline{d})^2
\end{equation}

Here, $\overline{d}$ is the average degree of the nodes in that graph. Across multiple graphs, the average degree variance is the mean of the degree variances of each graph.

\subsection{Summary} 
In our dataset, the directed graph has two categories: sparse and planar. The undirected graph has five categories: sparse, planar, regular, dense and complete. We summarized the experimental results based on these categories:
\begin{itemize}
[leftmargin=0.6cm]
    \item Directed sparse graphs have a moderate number of nodes and edges. The average degree is relatively low, indicating sparsity. The degree variance suggests variability in node connectivity.
    \item Directed planar graphs have fewer nodes and edges compared to sparse graphs. The average degree and variance are low, indicating even sparser connectivity with less variation in node degree.
    \item Undirected sparse graphs have a similar number of nodes to directed sparse graphs but significantly fewer edges. The average degree is just below 2, with low degree variance, indicating very sparse and consistent connectivity.
    \item Undirected planar graphs are similar to undirected sparse graphs in terms of average degree and degree variance but have fewer nodes and edges. They maintain sparse and consistent connectivity.
    \item Undirected regular graphs have a higher average number of edges and degree compared to sparse and planar graphs, indicating denser connectivity. The low degree variance suggests uniformity in node connectivity.
    \item Undirected complete graphs have a very high number of edges, as expected. Each node is connected to every other node, resulting in the maximum possible average degree for the given number of nodes. The degree variance is zero, indicating perfect uniformity in connectivity.
\end{itemize}

\begin{figure*}[htbp]
    \centering
    \subfigure[Execution Time Results on Directed Graph]{
        \includegraphics[width=0.45\textwidth]{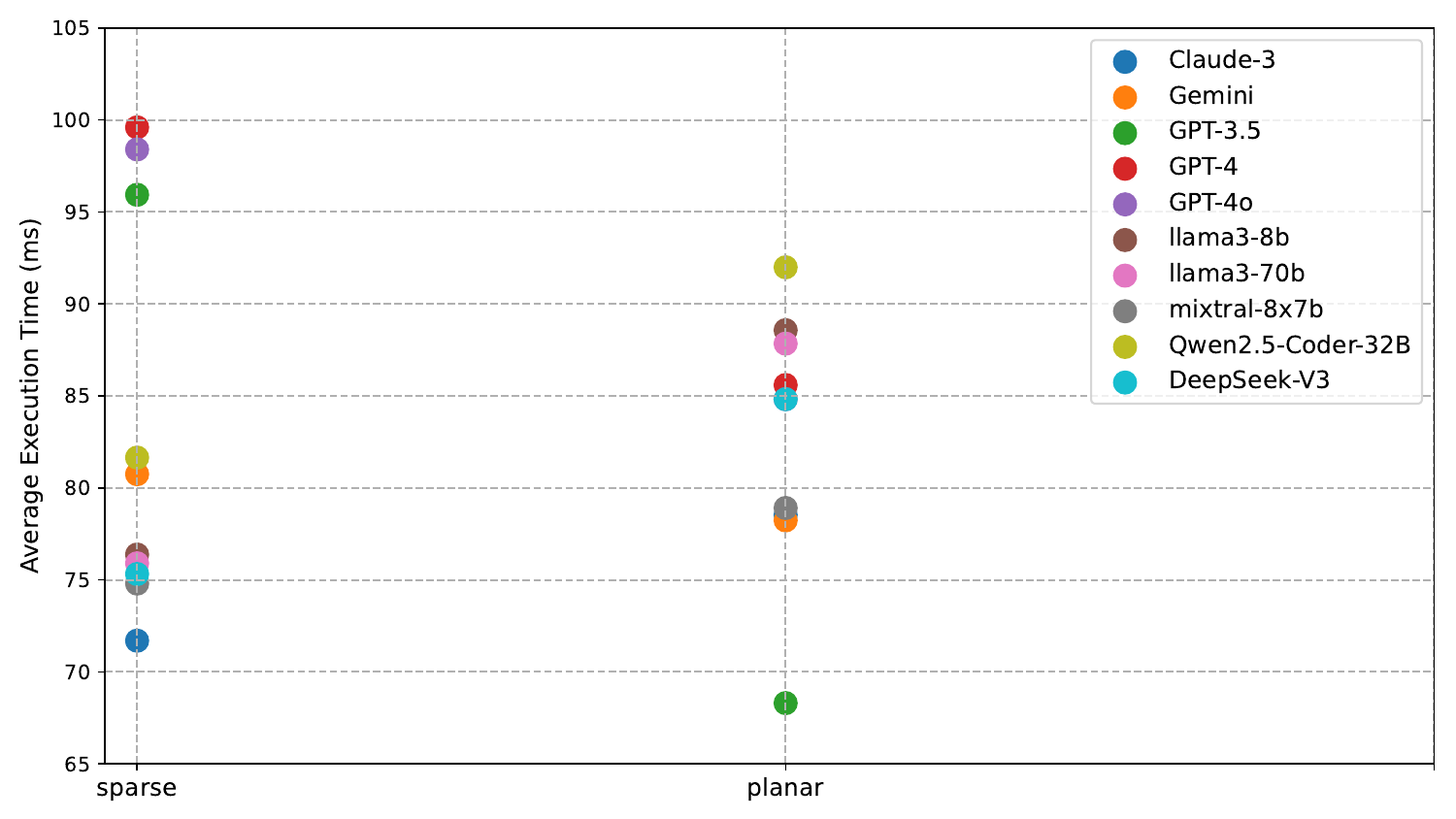}\label{fig:directed_graph_time}
    }
    \subfigure[Memory Usage on Directed Graph]{
        \includegraphics[width=0.45\textwidth]{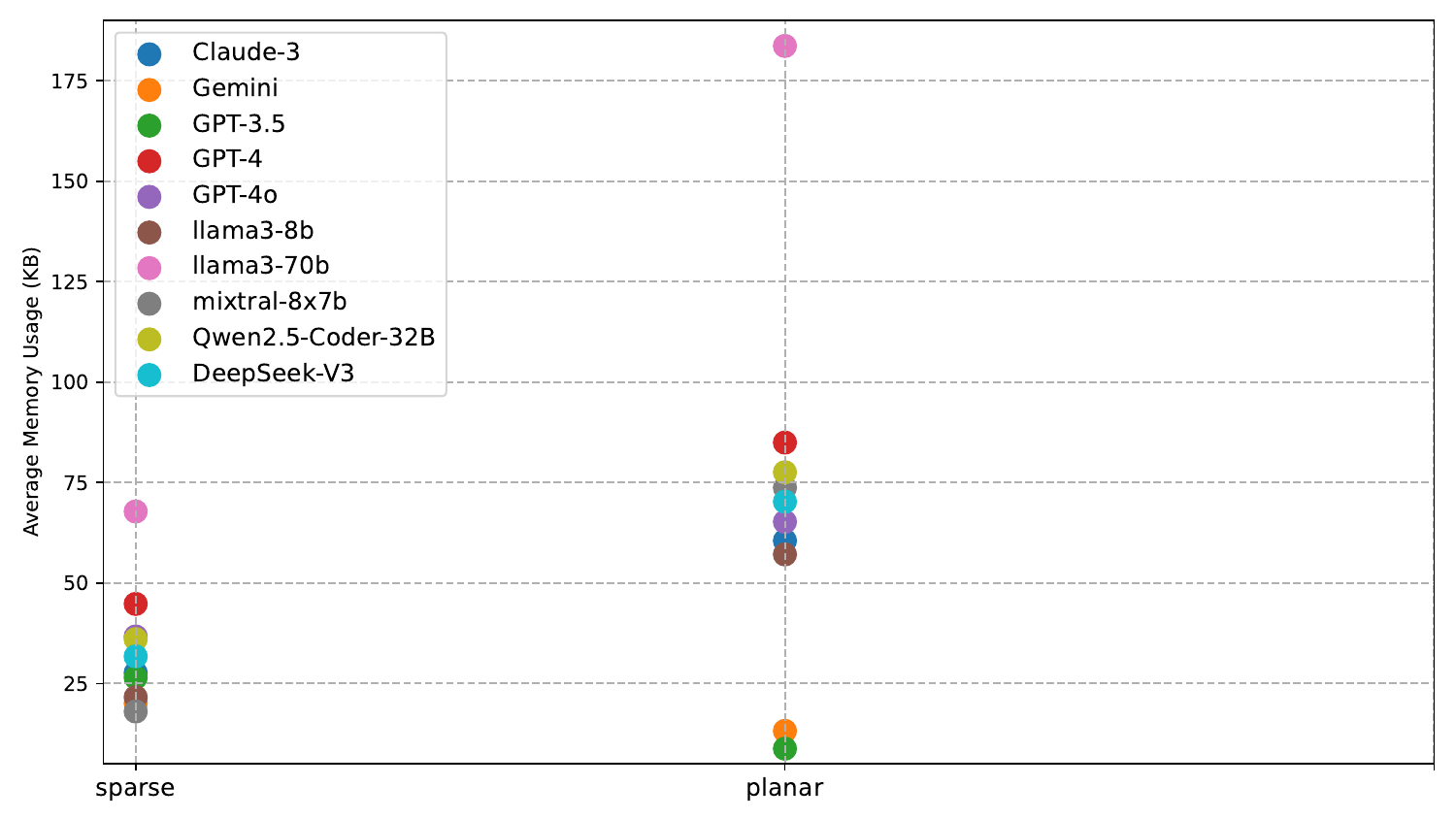}\label{fig:directed_graph_memory}
    }
    \subfigure[Execution Time Results on Undirected Graph]{
    \includegraphics[width=0.45\textwidth]{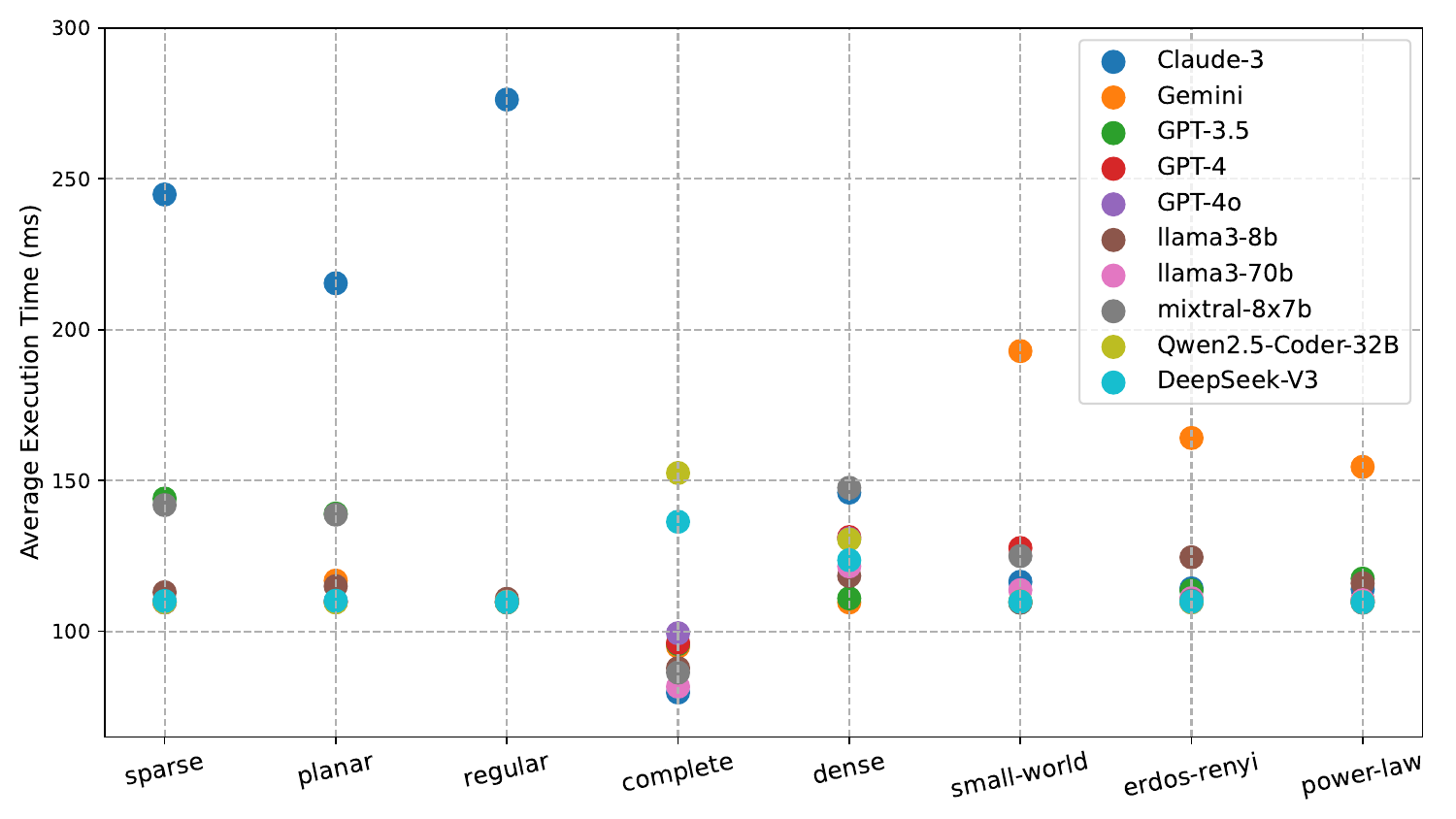}\label{fig:undirected_graph_time}
    }
    \subfigure[Memory Usage on Undirected Graph]{
    \includegraphics[width=0.45\textwidth]{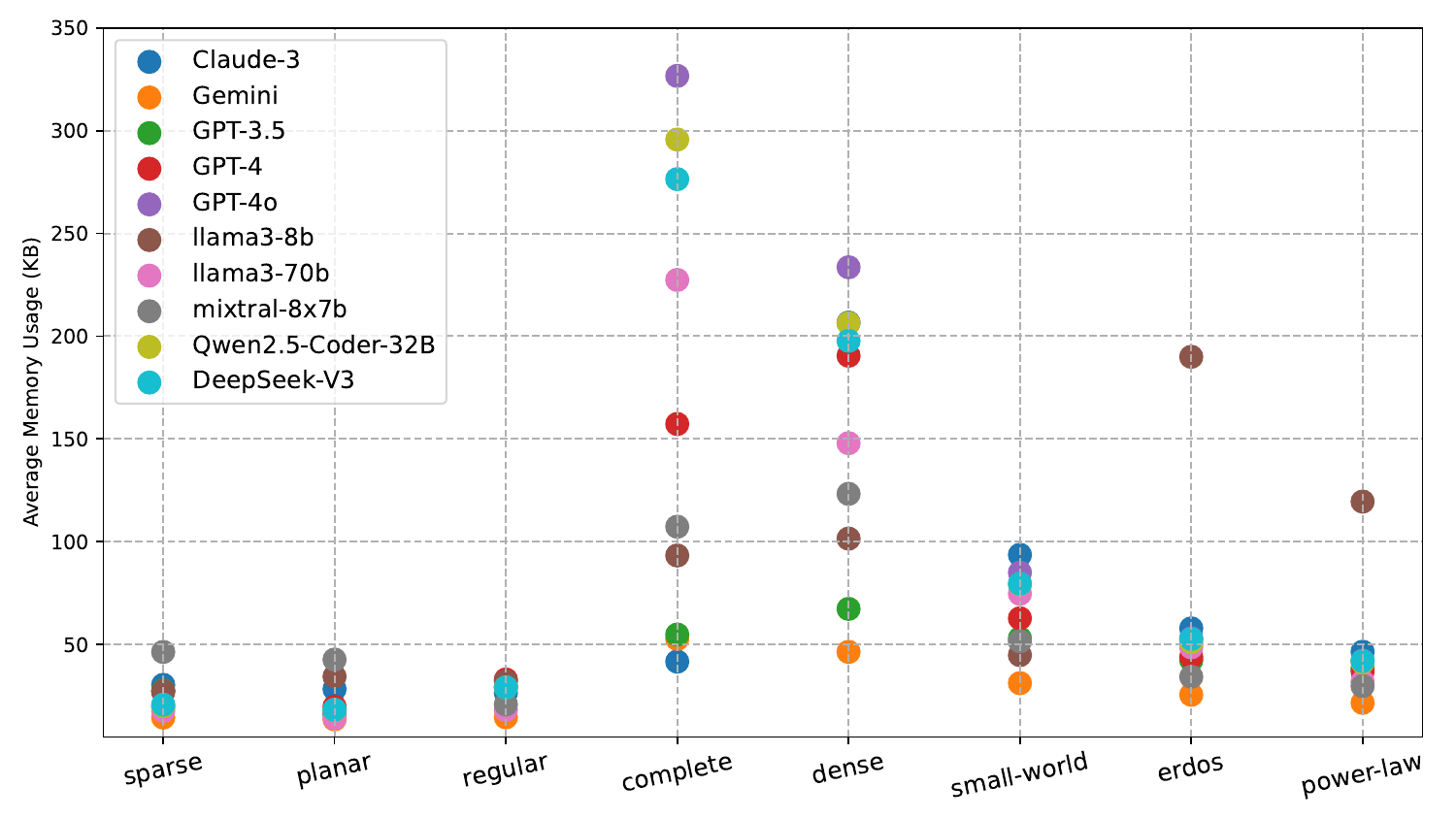}\label{fig:undirected_graph_memory}
    }
    \caption{Time and Memory usage results of LLMs on \datasetname.}\label{fig:time_and_usage}
\end{figure*}

\section{Time and Memory Usage}\label{sec:appendix-usage}
In addition to evaluating the passing rates of LLMs on our dataset, we also present the time and memory usage data, summarized in Figure \ref{fig:time_and_usage}. These metrics are crucial for assessing the efficiency of the code generated by LLMs.

\subsection{Results on Directed Graphs}
For directed graphs, GPT-3.5 stands out as the most efficient model, demonstrating the lowest average execution times and memory usage across planar and sparse graphs. Claude-3-sonnet also performs well, particularly for sparse graphs, achieving the lowest execution time. Gemini-pro and Mixtral-8x7b show good memory efficiency, with Mixtral-8x7b being particularly efficient for sparse graphs. Conversely, Llama-3-70b consistently exhibits the highest memory usage, making it the least efficient in terms of memory management. Although GPT-4 is powerful, it has higher execution times and memory usage compared to GPT-3.5. Both Qwen2.5-Coder-32B and DeepSeek-V3 show lower usage of time and memory, compared to other LLMs. Overall, GPT-3.5 emerges as the most balanced and efficient model across the tested scenarios, whereas Llama-3-70B exhibits notable inefficiencies, particularly in memory consumption. However, this analysis is based solely on execution time and memory usage; in practice, researchers should also consider model accuracy to obtain a more comprehensive evaluation.

\subsection{Results on Undirected Graphs}
The performance of models on undirected graphs varies significantly across different graph types. Claude-3-sonnet displays notably high execution times for planar, sparse, and regular graphs but performs exceptionally well with complete graphs, showing the lowest execution time albeit with relatively higher memory usage. Gemini demonstrates consistently low memory usage and good execution times, particularly excelling with sparse graphs. GPT-3.5 and GPT-4 show balanced performance, with GPT-3.5 achieving better execution times for planar graphs and GPT-4 managing memory more efficiently except for complete graphs, where it uses significantly more memory. GPT-4o is similar to GPT-4 but with slightly improved times and memory efficiency. Llama-3-8b and Llama-3-70b exhibit higher memory usage, especially for complete graphs, with Llama-3-70b showing the highest memory usage for complete graphs. Mixtral-8x7b presents high execution times and memory usage across all graph types but performs moderately well with regular graphs. For low-density graphs, Qwen2.5-Coder-32B and DeepSeek-V3 exhibit performance comparable to other LLMs. However, when processing more complex structures, such as complete graphs, both models maintain high execution time and memory consumption. Overall, Gemini and GPT-3.5 emerge as the most efficient models in terms of execution time and memory usage across various graph types, while Claude-3-sonnet and GPT-4 show significant variations based on the graph type.

It is important to note that high efficiency in code execution does not always guarantee a high passing rate. Occasionally, code may produce incorrect answers, which can result in accelerated program execution.

\section{LLM Performance on Directed and Unidirected Graphs}\label{sec:appendix-llm-d&und}
In the experiment section, we found that the passing rate of directed graphs is generally higher than that of undirected graphs. However, it is difficult to make a definitive statement since the Leetcode problems on directed and undirected graphs may not be at a similar level of difficulty. Also, we did not find analogous problems involving undirected graphs and directed graphs within Leetcode and so it is difficult to compare the performance of LLMs on directed and undirected graphs. We compute the statistical significance of the gap between directed graphs and undirected graphs (assuming that the ensemble of problems had a similar level of difficulty). We found that across most categories and models, there were statistically significant differences in the performance outcomes. 

The p-value results in Table \ref{tab:directed_vs_undirected} reveal notable differences in the performance of LLMs when solving graph-related problems on directed versus undirected graph samples. Among the eight models evaluated, Claude-3, Gemini, GPT-3.5, Llama-3-8b, and Mixtral-8x7b demonstrate statistically significant differences, with p-values well below the conventional threshold of 0.05. This suggests that these models exhibit distinct behavior or performance when handling directed and undirected graphs. In contrast, GPT-4, GPT-4o, and Llama-3-70b show no statistically significant difference, as indicated by their higher p-values (all above 0.05), implying that these models handle directed and undirected graph structures similarly. The variance in statistical significance across models may point to differing architectural strengths or limitations in their ability to generalize graph-based reasoning across different types of graph inputs. Overall, these results highlight the need for further investigation into model-specific behavior on varying graph structures, as some LLMs clearly struggle more than others with the distinction between directed and undirected graphs. For completeness, we also show examples of LLM reasoning on directed and undirected graphs in Section \ref{paragraph: LLM_performance_on_directed&undirected_graphs}.

\begin{table}[htbp]
\centering
\setlength{\tabcolsep}{2.5pt}

\begin{tabular}{lcc}
\hline
\textbf{Model} & \textbf{P-value} & \textbf{Statistical Significance} \\ \hline
Claude-3-sonnet       & 3.23e-6          & Significant                       \\
Gemini-pro         & 2.16e-2          & Significant                       \\
GPT-3.5        & 7.56e-3          & Significant                       \\
GPT-4          & 8.28e-1          & Not Significant                   \\
GPT-4o         & 8.02e-1          & Not Significant                   \\
Llama-3-8b   & 1.36e-2          & Significant                       \\
Llama-3-70b  & 6.93e-1          & Not Significant                   \\
Mixtral-8x7b & 1.79e-3          & Significant                       \\ \hline
\end{tabular}
\caption{Statistical Results of Directed vs. Undirected Graphs for LLMs.}
\label{tab:directed_vs_undirected}
\end{table}

\subsection{Comparison with Traditional Graph Machine Learning (ML) methods} 
The objective of our work is to establish a benchmark that evaluates how well LLMs understand and reason about graph data structures, rather than comparing their performance with specialized graph machine learning models. Graph machine learning models (e.g., GNNs) are designed explicitly to solve graph-based problems and are optimized for such tasks. In contrast, our focus is on understanding the extent to which LLMs, which are not specifically trained on graph problems, can generalize their capabilities to this domain. Our target is to assess their broader applicability in scenarios where graph reasoning is required but where graph-specific models might not be available or practical.

Our framework tests LLMs with graph problem statements and generates a program to solve them. The correctness of the program serves as a direct indicator of the LLMs' understanding and reasoning abilities. Our evaluation is based on this criterion, which is different from the evaluation of specialized graph machine learning models. We show the capabilities of LLMs on graph-structure problems in Table \ref{tab:test_dataset}, and we also demonstrate the improvement of LLMs with \methodname{} in Table \ref{tab:ssd_on_concepts_and_3_new_graphs} and Table \ref{table:ssd_results}. These results provide insights into the reasoning capabilities of LLMs on graph problems and show that our work serves as a complementary effort to existing graph ML research by exploring a different aspect of LLM capabilities. For any problem instance, our characterization is binary (either the produced code solved the problem or not) as opposed to the statistical evaluation in graph ML. Table \ref{tab:comparison} expands further on this comparison.

\begin{table*}[ht]
\centering
\renewcommand{\arraystretch}{0.7}
\begin{tabular}{@{}p{2.8cm}|p{6cm}|p{6cm}@{}}
\toprule
\textbf{Feature} & \textbf{Our Work} & \textbf{Traditional ML Methods (e.g., GNNs)} \\
\midrule
\textbf{Input Type} & Language-based queries & Pure graph-structure data \\
\midrule
\textbf{Main Process} & \begin{tabular}[c]{@{}l@{}}1. Understand graph problem queries\\ 2. Generate code to solve the problem\\ 3. Evaluate graph reasoning via\\ 36k test cases\end{tabular} & \begin{tabular}[c]{@{}l@{}}1. Apply graph algorithms directly\\ 2. Analyze graph structures\end{tabular} \\
\midrule
\textbf{Understanding and Reasoning} & \begin{tabular}[c]{@{}l@{}}Uses pre-trained knowledge and\\ language understanding to interpret\\ \textbf{graph-related queries}\end{tabular} & \begin{tabular}[c]{@{}l@{}}Uses \textbf{graph topology} and properties \\ for algorithmic reasoning\end{tabular} \\
\midrule
\textbf{Key Strengths} & \textbf{Code generative capabilities}, natural language queries, and unstructured inputs & High accuracy and efficiency in specific graph tasks \\
\midrule
\textbf{Typical Applications} & \begin{tabular}[c]{@{}l@{}}Language-interactive graph data\\ analysis, automatic \textbf{code generation},\\ interpreting complex graph data queries\end{tabular} & \begin{tabular}[c]{@{}l@{}}Node classification,\\ graph classification, link prediction\end{tabular} \\
\bottomrule
\end{tabular}
\caption{Comparison between our work and traditional machine learning methods.}
\label{tab:comparison}
\end{table*}

\subsection{Statistical Analysis}\label{paragraph:p-value_for_3_new_graphs} We compute the p-value to evaluate whether there are statistically significant differences in the passing rates of eight LLMs when applied to three types of graph structures. Small-world, Erdos-Renyi, and Power-law. A p-value below 0.05 indicates that the difference in performance between the models in the compared data sets is statistically significant, meaning that such a difference is unlikely to be due to random variation.

The statistical analysis of the passing rate results reveals that the differences in model performance between Small-world and Erdos-Renyi graph samples (p-value = 0.833) as well as between Small-world and Power-law graph samples (p-value = 0.762) are not statistically significant. These p-values, being substantially greater than typical significance thresholds (e.g., 0.05), indicate that there is no compelling evidence to suggest a significant difference in the performance of LLMs across these graph types. This implies that, from the perspective of coding and reasoning tasks involving connected graphs, the LLMs evaluated demonstrate comparable abilities when presented with Small-world, Erdos-Renyi, and Power-law structures.

Such findings are indicative of the generalization capabilities of LLMs in handling different types of graph topologies, which are commonly seen in practical scenarios. Given the similarities in model performance across these graph types, it can be inferred that the LLMs possess a consistent level of competence in reasoning and generating code for graph problems, regardless of the specific structural properties represented by these categories. This consistency is a promising outcome, suggesting that LLMs are potentially versatile in solving graph-related challenges irrespective of the complexity or type of underlying graph distributions. However, further investigation is needed to determine whether this generalizability holds for more complex graph characteristics beyond connectivity or across more challenging graph-related problem domains.

\section{Evaluation on the LeetCode Platform} We selected a total of 17 graph coding problems from the LeetCode platform for evaluation, categorized into 3 easy-level, 5 medium-level, and 9 hard-level problems, with difficulty levels defined by LeetCode. The evaluation process involved incorporating the problem statement, data examples, and code framework into a prompt. The LLMs are tasked with generating a complete code based on this prompt. The generated code is subsequently tested on the LeetCode platform to assess its accuracy and performance. The summarized results of this assessment are presented in Figure \ref{fig:online_leetcode_result}:
\begin{itemize}
[leftmargin=0.6cm]
\item For easy-level problems, most LLMs successfully pass the directed graph problems. Interestingly, private LLMs (e.g., the GPT family, Gemini-pro, and Claude-3-sonnet) outperform open-source LLMs (e.g., the Llama family and Mixtral-8x7b) on undirected graph problems.

\item For medium-level problems, LLMs demonstrate a better understanding of directed graph problems compared to undirected ones. Open-source LLMs perform worse than private LLMs, with a passing rate lower than 50\%. However, private LLMs and open-source LLMs perform comparably on directed graph problems.

\item For hard-level problems, only GPT-4 and GPT-4o achieve a 50\% passing rate on directed graph problems and a 20\% passing rate on undirected ones. All other LLMs fail to pass these problems, indicating that the GPT model exhibits the strongest reasoning ability among the tested LLMs.
\end{itemize}

\section{\methodname{} Results on Difficulty Levels}\label{sec:appendix-ssd-results}
We evaluate three versions of GPT models (GPT-3.5, GPT-4, and GPT-4o) on different difficulty levels of graph problems (same as the settings in Figure~\ref{fig:online_leetcode_result}). We focus on the hard-level problems due to the near-perfect performance on easier tasks. 
We compare the performance of \methodname{} method with the CoT-based method. The results are shown in Table~\ref{table:ssd_results}. Our method outperforms the CoT-based method on three GPT models, improving the passing rate from 0\% to 11.11\% for GPT-3.5, and from 33.30\% to 66.67\% for GPT-4 and GPT-4o. The results confirm the effectiveness of \methodname{} in enhancing LLMs' graph-solving capabilities. We observe a smaller improvement for GPT-3.5, likely due to its smaller model size and limited knowledge, making it less effective on hard-level graph tasks.

\begin{table}[htbp]
    \centering

    \resizebox{0.46\textwidth}{!}{
    \begin{tabular}{@{}c|ccc@{}}
    \toprule
    & \textbf{GPT-3.5} & \textbf{GPT-4} & \textbf{GPT-4o} \\ 
    \midrule
    \textbf{Easy-level (3)} & 100.00\% & 100.00\% & 100.00\% \\
    \textbf{Medium-level (5)} & 80\% & 100.00\% & 100.00\% \\
    \textbf{Hard-level (9)} & 0\% & 33.30\% & 33.30\% \\
    \textbf{Hard-level (9) + SSD} & 11.11\% & 66.67\% & 66.67\% \\
    \textbf{Performance Gain} & \textbf{+11.11\%} & \textbf{+33.37\%} & \textbf{+33.37\%} \\
    \bottomrule
    \end{tabular}
    }
    \captionof{table}{Passing rate comparison of GPT-3.5, GPT-4, and GPT-4o on graph problems, demonstrating LLM's graph reasoning can be enhanced by \methodname{}.}\label{table:ssd_results}
    \vspace{-1em}
\end{table}

\newpage
\input{code}

\clearpage
\section{Sample Problems under Different Graph Concepts}\label{sec:appendix-graph-concept}
In this section, we will present problem samples under each graph concept from our dataset. We have 9 concepts in total: Topological Sorting, Path Planning, Path Finding, Cycle Detection, Graph Traversal, Parent-Child Relations, Graph Construction, DAG Related Calculations and General Graph Problems. Note that we are only presenting the "problem statement" here.

\begin{tcolorbox}[
    colframe=black!50!white, colback=white, 
    boxrule=0.5mm, arc=0mm, width=\columnwidth, 
    fontupper=\fontsize{8pt}{1pt}\selectfont,
    coltitle=black,
    fonttitle=\bfseries,
    left=1mm, right=1mm,top=1mm, bottom=1mm,]

    \textbf{Concept:} \textit{Topological Sorting}
    \\
    
    \textbf{Problem Statement:} 
    \\
    
    \textit{You are given an integer n, which indicates that there are n courses labeled from 1 to n. You are also given a 2D integer array relations where relations[j] = [prevCoursej, nextCoursej] denotes that course prevCoursej has to be completed before course nextCoursej (prerequisite relationship). Furthermore, you are given a 0-indexed integer array time where time[i] denotes how many months it takes to complete the (i+1)th course. You must find the minimum number of months needed to complete all the courses following these rules: You may start taking a course at any time if the prerequisites are met. Any number of courses can be taken at the same time.Return the minimum number of months needed to complete all the courses. Note: The test cases are generated such that it is possible to complete every course (i.e., the graph is a directed acyclic graph)}
\end{tcolorbox}

\begin{tcolorbox}[
    colframe=black!50!white, colback=white, 
    boxrule=0.5mm, arc=0mm, width=\columnwidth, 
    fontupper=\fontsize{8pt}{1pt}\selectfont,
    coltitle=black,
    fonttitle=\bfseries,
    left=1mm, right=1mm,top=1mm, bottom=1mm,]

    \textbf{Concept:} \textit{Path Planning}
    \\
    
    \textbf{Problem Statement:} 
    \\
    
    \textit{You are given a directed graph of n nodes numbered from 0 to n - 1, where each node has at most one outgoing edge. The graph is represented with a given 0-indexed array edges of size n, indicating that there is a directed edge from node i to node edges[i]. If there is no outgoing edge from node i, then edges[i] == -1. Return the length of the longest cycle in the graph. If no cycle exists, return -1. A cycle is a path that starts and ends at the same node.}
\end{tcolorbox}

\begin{tcolorbox}[
    colframe=black!50!white, colback=white, 
    boxrule=0.5mm, arc=0mm, width=\columnwidth, 
    fontupper=\fontsize{8pt}{1pt}\selectfont,
    coltitle=black,
    fonttitle=\bfseries,
    left=1mm, right=1mm,top=1mm, bottom=1mm,]

    \textbf{Concept:} \textit{Path Finding}
    \\
    
    \textbf{Problem Statement:} 
    \\
    
    \textit{There is a bi-directional graph with n vertices, where each vertex is labeled from 0 to n - 1 (inclusive). The edges in the graph are represented as a 2D integer array edges, where each edges[i] = [ui, vi] denotes a bi-directional edge between vertex ui and vertex vi. Every vertex pair is connected by at most one edge, and no vertex has an edge to itself. You want to determine if there is a valid path that exists from vertex source to vertex destination. Given edges and the integers n, source, and destination, return true if there is a valid path from source to destination, or false otherwise.}
\end{tcolorbox}

\begin{tcolorbox}[
    colframe=black!50!white, colback=white, 
    boxrule=0.5mm, arc=0mm, width=\columnwidth, 
    fontupper=\fontsize{8pt}{1pt}\selectfont,
    coltitle=black,
    fonttitle=\bfseries,
    left=1mm, right=1mm,top=1mm, bottom=1mm,]

    \textbf{Concept:} \textit{Cycle Detection}
    \\
    
    \textbf{Problem Statement:} 
    \\
    
    \textit{You are given a positive integer n representing the number of nodes in an undirected graph. The nodes are labeled from 1 to n. You are also given a 2D integer array edges, where edges[i] = [ai, bi] indicates that there is a bidirectional edge between nodes ai and bi. Notice that the given graph may be disconnected. Divide the nodes of the graph into m groups (1-indexed) such that: Each node in the graph belongs to exactly one group. For every pair of nodes in the graph that are connected by an edge [ai, bi], if ai belongs to the group with index x, and bi belongs to the group with index y, then |y - x| = 1. Return the maximum number of groups (i.e., maximum m) into which you can divide the nodes. Return -1 if it is impossible to group the nodes with the given conditions.}
\end{tcolorbox}

\begin{tcolorbox}[
    colframe=black!50!white, colback=white, 
    boxrule=0.5mm, arc=0mm, width=\columnwidth, 
    fontupper=\fontsize{8pt}{1pt}\selectfont,
    coltitle=black,
    fonttitle=\bfseries,
    left=1mm, right=1mm,top=1mm, bottom=1mm,]

    \textbf{Concept:} \textit{Graph Traversal}
    \\
    
    \textbf{Problem Statement:} 
    \\
    
    \textit{There is an undirected graph with n nodes, numbered from 0 to n - 1. You are given a 0-indexed integer array scores of length n where scores[i] denotes the score of node i. You are also given a 2D integer array edges where edges[i] = [ai, bi] denotes that there exists an undirected edge connecting nodes ai and bi. A node sequence is valid if it meets the following conditions: There is an edge connecting every pair of adjacent nodes in the sequence. No node appears more than once in the sequence. The score of a node sequence is defined as the sum of the scores of the nodes in the sequence. Return the maximum score of a valid node sequence with a length of 4. If no such sequence exists, return -1.}
\end{tcolorbox}

\begin{tcolorbox}[
    colframe=black!50!white, colback=white, 
    boxrule=0.5mm, arc=0mm, width=\columnwidth, 
    fontupper=\fontsize{8pt}{1pt}\selectfont,
    coltitle=black,
    fonttitle=\bfseries,
    left=1mm, right=1mm,top=1mm, bottom=1mm,]

    \textbf{Concept:} \textit{Parent-Child Relations}
    \\
    
    \textbf{Problem Statement:} 
    \\
    
    \textit{You are given an integer n denoting the number of cities in a country. The cities are numbered from 0 to n - 1. You are also given a 2D integer array roads where roads[i] = [ai, bi] denotes that there exists a bidirectional road connecting cities ai and bi. You need to assign each city with an integer value from 1 to n, where each value can only be used once. The importance of a road is then defined as the sum of the values of the two cities it connects. Return the maximum total importance of all roads possible after assigning the values optimally.}
\end{tcolorbox}

\begin{tcolorbox}[
    colframe=black!50!white, colback=white, 
    boxrule=0.5mm, arc=0mm, width=\columnwidth, 
    fontupper=\fontsize{8pt}{1pt}\selectfont,
    coltitle=black,
    fonttitle=\bfseries,
    left=1mm, right=1mm,top=1mm, bottom=1mm,]

    \textbf{Concept:} \textit{Graph Construction}
    \\
    
    \textbf{Problem Statement:} 
    \\
    
    \textit{You are given an array pairs, where pairs[i] = [xi, yi], and there are no duplicates. xi < yi. Let ways be the number of rooted trees that satisfy the following conditions: The tree consists of nodes whose values appeared in pairs. A pair [xi, yi] exists in pairs if and only if xi is an ancestor of yi or yi is an ancestor of xi. Note: the tree does not have to be a binary tree. Two ways are considered to be different if there is at least one node that has different parents in both ways. Return 0 if ways equal to 0. Return 1 if ways equal to 1. Return 2 if ways are larger than 1. A rooted tree is a tree that has a single root node, and all edges are oriented to be outgoing from the root. An ancestor of a node is any node on the path from the root to that node (excluding the node itself). The root has no ancestors.}
\end{tcolorbox}

\begin{tcolorbox}[
    colframe=black!50!white, colback=white, 
    boxrule=0.5mm, arc=0mm, width=\columnwidth, 
    fontupper=\fontsize{8pt}{1pt}\selectfont,
    coltitle=black,
    fonttitle=\bfseries,
    left=1mm, right=1mm,top=1mm, bottom=1mm,]

    \textbf{Concept:} \textit{DAG Related Calculations}
    \\
    
    \textbf{Problem Statement:} 
    \\
    
    \textit{You are given a directed graph of n nodes numbered from 0 to n - 1, where each node has at most one outgoing edge. The graph is represented with a given 0-indexed array edges of size n, indicating that there is a directed edge from node i to node edges[i]. If there is no outgoing edge from node i, then edges[i] == -1. Return the length of the longest cycle in the graph. If no cycle exists, return -1. A cycle is a path that starts and ends at the same node.}
\end{tcolorbox}

\begin{tcolorbox}[
    colframe=black!50!white, colback=white, 
    boxrule=0.5mm, arc=0mm, width=\columnwidth, 
    fontupper=\fontsize{8pt}{1pt}\selectfont,
    coltitle=black,
    fonttitle=\bfseries,
    left=1mm, right=1mm,top=1mm, bottom=1mm,]

    \textbf{Concept:} \textit{General Graph Problems}
    \\
    
    \textbf{Problem Statement:} 
    \\
    
    \textit{There is an undirected graph consisting of n nodes numbered from 1 to n. You are given the integer n and a 2D array edges where edges[i] = [ai, bi] indicates that there is an edge between nodes ai and bi. The graph can be disconnected. You can add at most two additional edges (possibly none) to this graph so that there are no repeated edges and no self-loops. Return true if it is possible to make the degree of each node in the graph even, otherwise return false. The degree of a node is the number of edges connected to it.}
\end{tcolorbox}

\clearpage
\section{An Example of How LLMs Solve Difficult Problems}\label{sec:appendix-graph-concept-solving}
In this section, we will present how do LLMs solve a difficult concept problem. We choose Claude-3, Gemini, GPT-4 and GPT-4o to be verification models. First, we will present the chosen problem. And then, the reasoning and coding process of LLMs will be shown.

\definecolor{l_yellow}{HTML}{c8b372}

\begin{tcolorbox}[colback=white, colframe=l_yellow, title=Graph Question - Concept: Path Finding,
coltitle=black, 
fontupper=\fontsize{7pt}{2pt}\selectfont, ]

\textbf{[Problem Statement]}

You are given three positive integers n, x, and y. In a city, there exist houses numbered 1 to n connected by n streets. There is a street connecting the house numbered i with the house numbered i + 1 for all 1 <= i <= n - 1 . An additional street connects the house numbered x with the house numbered y. For each k, such that 1 <= k <= n, you need to find the number of pairs of houses (house1, house2) such that the minimum number of streets that need to be traveled to reach house2 from house1 is k. Return a 1-indexed array result of length n where result[k] represents the total number of pairs of houses such that the minimum streets required to reach one house from the other is k. Note that x and y can be equal.
\\

\textbf{[Data Examples]}

Input: n = 3, x = 1, y = 3
Output: [6,0,0]

Input: n = 5, x = 2, y = 4
Output: [10,8,2,0,0]

\end{tcolorbox}

\definecolor{green}{HTML}{B7C47D}

\begin{tcolorbox}[colback=white, colframe=green, title=GPT-4o Solution, coltitle=black, fontupper=\fontsize{7pt}{2pt}\selectfont]

Steps to solve:\\
1. Initialize result array: Start with an array of zeroes to store the count of pairs.\\
2. Iterate through pairs: For each pair (i, j), compute both the direct distance and the alternative path distance using the extra street between x and y.\\
3. Calculate min distance: Find the minimum of the two distances and update the corresponding index in the result array.\\
4. Return the result: Once all pairs are processed, return the result array.
\end{tcolorbox}

\begin{lstlisting}[caption={Code 5: Code generated by the GPT-4o.}]
class Solution:
    def countOfPairs(self, n: int, x: int, y: int) -> List[int]:
        # Initialize the result array with n zeroes
        result = [0] * n
        
        # Adjust x and y for 0-based indexing
        x -= 1
        y -= 1
        
        # Iterate over all possible pairs of houses (i, j)
        for i in range(n):
            for j in range(i + 1, n):
                # Direct distance between i and j
                direct_dist = j - i
                
                # Distance via the special street between x and y
                via_x_y = abs(i - x) + 1 + abs(j - y)
                via_y_x = abs(i - y) + 1 + abs(j - x)
                
                # Minimum distance between the direct path and the path via x and y
                min_dist = min(direct_dist, via_x_y, via_y_x)
                
                # Increment the count for the corresponding distance
                result[min_dist] += 1
        
        return result
\end{lstlisting}

\begin{tcolorbox}[colback=white, colframe=green, title=GPT-4 Solution, coltitle=black, fontupper=\fontsize{7pt}{2pt}\selectfont]

Steps to solve:\\
1. Construct the Graph: We'll use a simple adjacency list to represent the graph.\\
2. Shortest Path Calculation: For small values of n, a simple Floyd-Warshall algorithm can be considered since it computes shortest paths for all pairs. But the time complexity is too high. Given the specific graph structure (almost a path plus one edge), BFS from each node could be more efficient, leveraging the linear structure for faster computation.\\
3. Count the Pairs: After determining the shortest path distances, count the number of pairs for each distance.
\end{tcolorbox}

\begin{lstlisting}[caption={Code 6: Code generated by the GPT-4.}]
class Solution:
    def countOfPairs(self, n: int, x: int, y: int) -> List[int]:
        # Create the graph
        graph = defaultdict(list)
        for i in range(1, n):
            graph[i].append(i + 1)
            graph[i + 1].append(i)
        if x != y:  # Add the additional connection if x and y are not the same
            graph[x].append(y)
            graph[y].append(x)

        # Function to calculate distances using BFS
        def bfs(start):
            distances = [-1] * (n + 1)
            queue = deque([start])
            distances[start] = 0
            while queue:
                current = queue.popleft()
                current_dist = distances[current]
                for neighbor in graph[current]:
                    if distances[neighbor] == -1:
                        distances[neighbor] = current_dist + 1
                        queue.append(neighbor)
            return distances
        
        # Distance array where result[k] counts the pairs of distance k
        result = [0] * (n + 1)  # Using n+1 for convenience, will slice later

        # Calculate distances for each house and count the pairs
        for i in range(1, n + 1):
            distances = bfs(i)
            for j in range(i + 1, n + 1):  # Only count each pair once
                if distances[j] > 0:
                    result[distances[j]] += 1

        return result[1:n]  # We return the list from 1 to n-1 as specified
\end{lstlisting}


\begin{tcolorbox}[colback=white, colframe=green, title=Gemini Solution, coltitle=black, fontupper=\fontsize{7pt}{2pt}\selectfont]

Steps to solve:\\
1. Graph Representation: Create an adjacency list to represent the graph, where each node represents a house and edges connect adjacent houses.\\
2. BFS for Distances: Perform breadth-first search (BFS) starting from house x to calculate the shortest distances to all other houses.\\
3. Distance-Based Counting: Iterate through all pairs of houses and count the number of pairs where the sum of their distances is less than n (to avoid counting pairs that wrap around the circle).\\
4. Result Array: Store the counts in the result array, where result[k] represents the number of pairs with a distance of k.

\end{tcolorbox}

\begin{lstlisting}[caption={Code 7: Code generated by the Gemini.}]
class Solution:
    def countOfPairs(self, n: int, x: int, y: int) -> List[int]:
        # Create a graph representation using an adjacency list
        graph = [[] for _ in range(n)]
        graph[x - 1].append(y - 1)
        graph[y - 1].append(x - 1)

        # Perform breadth-first search (BFS) to calculate distances
        queue = [(0, x - 1)]  # Start with distance 0 from house x
        visited = set()
        distances = [0] * n
        while queue:
            distance, node = queue.pop(0)
            if node not in visited:
                visited.add(node)
                distances[node] = distance
                for neighbor in graph[node]:
                    if neighbor not in visited:
                        queue.append((distance + 1, neighbor))

        # Count pairs based on distances
        result = [0] * n
        for i in range(1, n):
            for j in range(i + 1, n):
                distance = distances[i - 1] + distances[j - 1]
                if distance < n:
                    result[distance] += 1

        return result
\end{lstlisting}

\begin{tcolorbox}[colback=white, colframe=green, title=Claude-3 Solution, coltitle=black, fontupper=\fontsize{7pt}{2pt}\selectfont]

Steps to solve:\\
1. We start by initializing a distance matrix dist with dimensions n x n, setting all initial distances to infinity.\\
2. We set the distances for adjacent houses (i and i+1) to 1, and the distance from each house to itself to 0. We also set the special connection between houses x and y to 1.\\
3. We use the Floyd-Warshall algorithm to compute the shortest paths between all pairs of houses. This algorithm has a time complexity of O($n^3$), but it's efficient for this problem size and gives us all the information we need.\\
4. After computing all shortest paths, we count the number of pairs for each distance. We iterate through all pairs of houses (i, j) where i < j, and increment the count for the corresponding distance. We increment by 2 to account for both (i, j) and (j, i) pairs.\\
5. Finally, we return the count array, which represents the number of pairs for each distance from 1 to n.

\end{tcolorbox}

\begin{lstlisting}[caption={Code 8: Code generated by the Claude-3.}]
class Solution:
    def countOfPairs(self, n: int, x: int, y: int) -> List[int]:
        # Initialize the distance matrix
        dist = [[float('inf')] * n for _ in range(n)]
        
        # Set distances for adjacent houses and the special connection
        for i in range(n):
            if i > 0:
                dist[i][i-1] = dist[i-1][i] = 1
            dist[i][i] = 0
        dist[x-1][y-1] = dist[y-1][x-1] = 1
        
        # Floyd-Warshall algorithm to compute all-pairs shortest paths
        for k in range(n):
            for i in range(n):
                for j in range(n):
                    dist[i][j] = min(dist[i][j], dist[i][k] + dist[k][j])
        
        # Count pairs for each distance
        count = [0] * n
        for i in range(n):
            for j in range(i+1, n):
                d = dist[i][j]
                if d < n:
                    count[d-1] += 2  # Count both (i,j) and (j,i)
        
        return count
\end{lstlisting}

\end{document}

%% file: intro.tex

\section{Introduction}

\begin{figure*}[htb]
    \centering\includegraphics[width=0.9\linewidth]{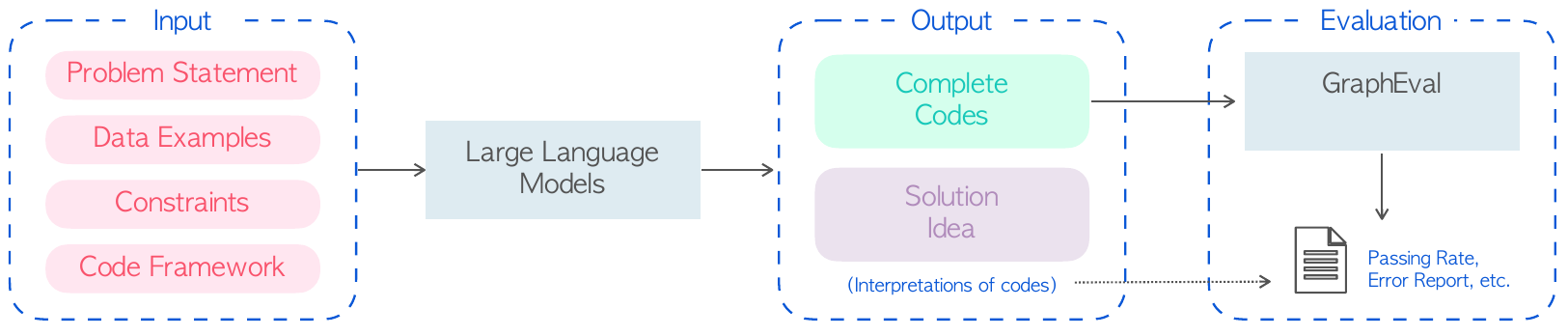}\caption{Overview of the Evaluation Framework. For each problem, we input problem statement, data examples, and code framework to LLMs. The LLMs generate the corresponding code and provide explanations. Finally, we evaluate the code on \datasetname{} and return the score details.}
    \label{fig:pipeline}
\end{figure*}

Large language models (LLMs) such as GPTs~\citep{achiam2023gpt,brown2020language,chen2021evaluating}, Gemini~\citep{team2023gemini,reid2024gemini}, Claude-3~\citep{claude3}, LLaMA-3~\citep{touvron2023llama}, Mixtral~\citep{jiang2024mixtral}, DeepSeek-V3~\citep{liu2024deepseek} and Qwen-2.5-coder\citep{hui2024qwen2} have achieved remarkable success in solving a wide range of natural language processing (NLP) tasks: for example, question answering~\citep{devlin2018bert, gpt3, raffel2020exploring}), machine translation~\citep{raffel2020exploring, gpt3}, text classification~\citep{raffel2020exploring, yang2019xlnet, liu2019roberta}, and text generation~\citep{yang2019xlnet, achiam2023gpt}.
However, LLMs have difficulty with complex coding problems, particularly those involving structured data like graphs~\citep{zhang2023graph}.

Current research highlights that while LLMs can handle basic graph-related queries, their performance declines on more complex coding challenges involving graph algorithms and multi-step problem solving~\citep{liu2023evaluating,wang2024can,creswell2022selection}. These shortcomings highlight the need for targeted evaluation and improvement of LLM's coding abilities in graph-related tasks~\citep{liu2024codexgraph,cai2024codegraph}.


To analyze the above gaps, we propose \datasetname{}, the \textbf{first dataset} designed to evaluate the graph-solving capabilities of LLMs through coding problems. \datasetname{} includes 40 graph coding problems with 36,900 test cases, covering a wide range of graph characteristics and algorithmic challenges. Each problem provides: (1) a problem statement, (2) data examples, (3) constraints, and (4) a coding framework for LLMs to build solutions. The dataset is organized into eight primary categories of graph structures: sparse, planar, regular, dense, complete~\citep{diestel2024graph}, Small-world~\citep{watts1998collective}, Erdos-Renyi~\citep{erdos1960evolution}, and Power-law~\citep{barabasi1999emergence} graphs, with sub-categories such as connected, disconnected, cyclic, and acyclic graphs to ensure comprehensive coverage. Based on \datasetname{}, we propose an evaluation framework that is designed to give immediate feedback by returning failed test cases and execution details, promoting deeper model understanding and troubleshooting. This mechanism differentiates our approach from traditional coding platforms (e.g., LeetCode), where test case details are often hidden from users~\citep{hou2024systematic, hu2024leveraging}.

To further enhance the usability of our evaluation framework and \datasetname{}, we propose an instruction-based method, Structured Symbolic Decomposition (\methodname{}). Inspired by human problem-solving techniques~\citep{paas2020cognitive}, \methodname{} decomposes complex tasks into manageable components: a ``cognitive step'' for understanding the problem and an ``action step'' for implementing the solution. The experiments show that \methodname{} enhances the average passing rate of GPT-4, GPT-4o, Gemini-Pro and Claude-3 Sonnet by 8.38\%, 6.78\%, 29.28\% and 25.28\%, respectively.


Figure~\ref{fig:pipeline} presents an overview of our evaluation framework that consists of three steps: problem selection, code generation, and evaluation. 
Our contributions are summarized as follows:
\begin{enumerate}[leftmargin=0.6cm]
    \item We introduce \datasetname{}, a graph coding and benchmarking dataset with 40 coding problems and 36,900 test cases, designed to evaluate LLMs' graph-solving abilities across diverse and complex graphs.
    
    \item We evaluate the performance of ten LLMs across various graph types (directed vs. undirected, Power-law, Small-world, Erdos-Renyi), fundamental graph concepts (traversal, construction, path finding, topological sorting, cycle detection), and problem difficulty levels (easy, medium, hard). 
    \item We propose \methodname{}, an instruction-based method that decomposes complex problems into reasoning components for LLMs. Experiments show \methodname{} improves performance by an average of 17.43\% across GPT-4, GPT-4o, Gemini-Pro, and Claude-3-Sonnet.
    \item We develop an evaluation framework with real-time feedback, paired with \methodname{} to improve LLM performance on complex graph tasks, particularly for models with lower baseline performance, yielding up to a 48.50\% improvement.

\end{enumerate}

Our work examines LLM performance on graph problems by categorizing difficulty levels, fundamental concepts, and graph types while evaluating state-of-the-art LLMs. We provide insights into LLM capabilities in graph coding and manipulation, offering a roadmap for future advancements. 
For LLM users, our work clarifies when and how LLMs can effectively solve graph problems, highlighting their strengths and limitations. Additionally, we introduce \methodname{} to improve LLM performance on complex graph tasks.

%% file: table2.tex
\begin{table*}[htb]
    \centering
    \resizebox{\textwidth}{!}{%
    \begin{tabular}{l l p{3.5cm} p{3.5cm} p{3cm} c}
        \toprule
        \textbf{Dataset Name} & \textbf{Graph Problems} & \textbf{Reasoning Scope} & \textbf{Graph Types} & \textbf{Scale} & \textbf{LLM Benchmark} \\
        \midrule
        CLRS-30 ~\citep{velivckovic2022clrs} & Algorithmic & Limited & Planar, Trees & 30 Problems & No \\
        BIG-Bench ~\citep{srivastava2022beyond} & Logic Puzzles & Indirect & Small Graphs & 204 Tasks & Yes \\
        ProofWriter ~\citep{tafjord2021proofwriter} & Logical Reasoning & Limited & Implicit graphs & 16k Examples & Yes \\
        PrOntoQA ~\citep{misra2023prontoqa} & Ontology Reasoning & Specific & Taxonomies & 80k QA pairs & Yes \\
        GraCoRe ~\citep{yuan2024gracore} & Graph Reasoning & Pure \& Heterogeneous Graphs & --- & 5k Graphs & Yes \\
        HGB ~\citep{li2023hybrid} & Hybrid Graph Analysis & Complex Structures & Biology, Social & 23 Datasets & No \\
        
        \midrule
        
        \datasetname{} (\textbf{Ours}) & Graph Coding and Reasoning & Comprehensive & 8 Main Categories with 4 Sub-categories & 40 Problems with 36.9k Cases & Yes \\
        \bottomrule
    \end{tabular}%
    }\caption{Comparison of \datasetname{} with existing datasets for evaluating LLMs on graph-related tasks.}
    \label{tab:dataset_comparison}
\end{table*}

%% file: table1.tex
\begin{table*}[htb]
    \centering
    \resizebox{\textwidth}{!}{
    \renewcommand{\arraystretch}{0.4} 
    \begin{tabular}{@{}cc|cccccccccc@{}}
    \toprule
    \multicolumn{2}{c|}{}                                                  & \textbf{\begin{tabular}[c]{@{}c@{}}Claude-3\\ -Sonnet\end{tabular}} & \textbf{\begin{tabular}[c]{@{}c@{}}Gemini\\ -Pro\end{tabular}} & \textbf{\begin{tabular}[c]{@{}c@{}}GPT\\ -3.5\end{tabular}} & \textbf{\begin{tabular}[c]{@{}c@{}}GPT\\ -4\end{tabular}} & \textbf{\begin{tabular}[c]{@{}c@{}}GPT\\ -4o\end{tabular}} & \textbf{\begin{tabular}[c]{@{}c@{}}Llama-3\\ -8b\end{tabular}} & \textbf{\begin{tabular}[c]{@{}c@{}}Llama-3\\ -70b\end{tabular}} & \textbf{\begin{tabular}[c]{@{}c@{}}Mixtral\\ -8x7b\end{tabular}} & \textbf{\begin{tabular}[c]{@{}c@{}}Qwen2.5\\ -Coder-32B\end{tabular}} & \textbf{\begin{tabular}[c]{@{}c@{}}DeepSeek\\ -V3\end{tabular}} \\ \midrule
    \multirow{7}{*}{\centering \textbf{SG}} & c   & 68 | 39 & 55 | 41 & 69 | 48 & 61 | 66 & 64 | 77 & 46 | 27 & 59 | 70 & 81 | 57 & 74 | 72 & 61 | \textbf{92} \\ \cmidrule(lr){2-2}
    & dc  & 54 | 37 & 31 | 40 & 59 | 61 & 66 | 68 & \textbf{76} | 81 & 25 | 32 & 51 | 62 & 64 | 60 & 73 | 66 & 70 | \textbf{96} \\ \cmidrule(lr){2-2}
    & cy  & 69 | 48 & 46 | 32 & 64 | 71 & 50 | 80 & 59 | 88 & 42 | 40 & 67 | 74 & \textbf{73} | 64 & \textbf{73} | 79 & 67 | \textbf{97} \\ \cmidrule(lr){2-2}
    & acy & 63 | 50 & 41 | 38 & 75 | 48 & 68 | 59 & 78 | 69 & 36 | 25 & 51 | 58 & 68 | 53 & \textbf{84} | 64 & 79 | \textbf{89} \\ \midrule
    \multirow{7}{*}{\centering \textbf{PG}} & c   & \textbf{80} | 30 & 47 | 26 & 63 | 41 & 60 | 55 & 60 | 65 & 45 | 21 & 75 | 60 & 60 | 48 & 75 | 56 & 75 | \textbf{75} \\ \cmidrule(lr){2-2}
    & dc  & 64 | 39 & 39 | 37 & 61 | 47 & 80 | 70 & \textbf{88} | 69 & 29 | 20 & 52 | 44 & 65 | 56 & 86 | 62 & 80 | \textbf{89} \\ \cmidrule(lr){2-2}
    & cy  & \textbf{68} | 38 & 46 | 28 & 61 | 59 & 58 | 71 & 67 | 73 & 37 | 30 & 64 | 62 & 64 | 53 & \textbf{68} | 70 & 61 | \textbf{84} \\ \cmidrule(lr){2-2}
    & acy & 61 | 40 & 42 | 39 & 69 | 42 & 73 | 55 & 85 | 65 & 36 | 20 & 41 | 47 & 61 | 51 & \textbf{90} | 61 & 80 | \textbf{79} \\ \midrule
    \multirow{5}{*}{\centering \textbf{RG}} & c   & NA | 65 & NA | 39 & NA | 68 & NA | 85 & NA | \textbf{97} & NA | 53 & NA | 81 & NA | 75 & NA | 89 & NA | 95 \\ \cmidrule(lr){2-2}
    & dc  & NA | 61 & NA | 38 & NA | 58 & NA | 86 & NA | \textbf{94} & NA | 42 & NA | 79 & NA | 73 & NA | 77 & NA | 93 \\ \cmidrule(lr){2-2}
    & cy  & NA | 68 & NA | 50 & NA | 74 & NA | 87 & NA | \textbf{99} & NA | 56 & NA | 84 & NA | 77 & NA | 90 & NA | 97 \\ \midrule
    \textbf{DG} & c   & 38 | 60 & 32 | 36 & 37 | 60 & 48 | 75 & 48 | 81 & \textbf{49} | 27 & 37 | 70 & 39 | 46 & 27 | 65 & 20 | \textbf{82} \\ \midrule
    \textbf{CG} & c   & NA | 33 & NA | 38 & NA | 67 & NA | \textbf{86} & NA | 64 & NA | 43 & NA | 50 & NA | 57 & NA | 67 & NA | 68 \\ \bottomrule
    \end{tabular}
    }
    \caption{Evaluation Results on \datasetname. Passing rates (\%) of ten LLMs across graph categories. The first column categorizes graphs: ``SG'' (sparse), ``PG'' (planar), ``RG'' (regular), ``DG'' (dense), and ``CG'' (complete). Abbreviations include ``c'' (connected), ``dc'' (disconnected), ``cy'' (cyclic), and ``acy'' (acyclic). Results are shown as ``Directed | Undirected'' for each category, with ``NA'' indicating not applicable. Bold values highlight the highest passing rate per row.}
    \label{tab:test_dataset}
\end{table*}

%% file: method.tex
\section{Improving LLM Graph Solving}
To enhance the usability of our evaluation framework and \datasetname{}, we introduce Structured Symbolic Decomposition (\methodname{}), an instruction-based method utilizing test cases from \datasetname{} for graph problems. Our approach aims to enable LLMs to perform better graph problem-solving processes.

\paragraph{Methodology} We hypothesize that decomposing complex graph problems into smaller, more manageable sub-problems and turning them into symbolic forms~\citep{dinu2024symbolicai,fang2024large, yang2024leandojo} will enhance the graph-solving capabilities of LLMs. Current methods rely on implicit knowledge and lack explicit guidance~\citep{wei2022chain,jin2024graph,huang2024prodigy}, leading to suboptimal performance, especially in complex scenarios. 
Our method mirrors human cognitive strategies~\citep{paas2020cognitive,romero2023synergistic}, which simplify complex tasks by decomposing them into two parts: \textit{cognitive step} and \textit{action step}, thereby improving comprehension and facilitating more effective solutions. 
We selected the problems from \datasetname{} to be evaluation problems. The test cases are used for problem understanding and program testing.

\paragraph{Instructions for LLMs} The instructions are composed of four parts: problem clarification, problem breakdown, solution formulation, and program implementation.

\begin{itemize}[leftmargin=0.3cm]
    \item \textbf{Problem Clarification:} \\
        \textbf{Cognitive Step:} \textit{You must first understand and clearly articulate the problem, including all inputs and desired outputs.}\\
         \textbf{Action Step:} \textit{Identify and list any specific rules, constraints, or conditions that influence the solution. Use the \textbf{\{test\_case\}} examples to assist the understanding.}
        
    \item \textbf{Problem Breakdown:} \\
            \textbf{Cognitive Step:} \textit{Decompose the problem into smaller, manageable sub-problems, translating it into a symbolic form and identifying the key components and relationships within the problem.}\\
            \textbf{Action Step:} \textit{Outline the sequential steps required to solve the overall problem.}
        
    \item \textbf{Solution Formulation:}\\
            \textbf{Cognitive Step:} \textit{Formulate solving strategies using the symbolic form developed in the previous step and define the algorithms and methods needed to address each sub-problem.}\\
            \textbf{Action Step:}
            \textit{Detail the algorithms and approaches for each sub-problem, ensuring they are logically connected and comprehensive.}
    \item \textbf{Program Implementation:}\\
            \textbf{Cognitive Step:} \textit{Conceptualize the implementation of each solution component.}\\
            \textbf{Action Step:} \textit{Write a program for each breakdown part, ensuring it aligns with the formulated strategy. Run \textbf{\{test\_cases\}} to verify the correctness of each component.}
\end{itemize}

We demonstrate the comparison of \methodname{}-enhanced method with CoT-based graph solving in the following example. The demonstration involves a ``shortest path'' question and test cases from \datasetname{}:
\begin{tcolorbox}[
    colframe=black!50!white, colback=white, 
    boxrule=0.5mm, arc=0mm, width=\columnwidth, 
    fontupper=\fontsize{8pt}{1pt}\selectfont,
    coltitle=black,
    fonttitle=\bfseries,
    left=1mm, right=1mm,top=1mm, bottom=1mm,]
    
    \textbf{Problem Statement:} 
    \textit{Let \( G = (V, E) \) be a bi-directional graph with \( n \) vertices, labeled from \( 0 \) to \( n-1 \). The edges in the graph are given as a 2D integer array \textit{edges}, where each \textit{edges[i]} = \([ u_i, v_i ]\) represents an edge between vertex \( u_i \) and vertex \( v_i \). There is no self-loop, and at most one edge exists between any two vertices.
    Return the length of the shortest cycle in the graph. If no cycle exists, return \(-1\).
    A cycle is a path that starts and ends at the same vertex, using each edge only once.}
    
    \textbf{Test Cases:}
    
    \textbf{Input:}
    \textit{ 
    \{`edges': [[0, 1], [2, 4], [4, 0], [2, 1], [4, 1], [0, 2], [3, 2], [1, 3]]\}, 
    \{`edges': [[0, 1], [0, 4], [1, 5], [4, 2], [3, 0], [2, 3], [5, 3], [4, 1], [5, 2]]\}, 
    \{`edges': [[0, 1], [1, 2], [3, 1], [0, 3], [2, 3], [0, 2]]\}, 
    \{`edges': [[0, 1], [1, 2]]\}, 
    \{`edges': [[4, 2], [5, 1], [5, 0], [0, 3], [5, 2], [1, 4], [1, 3], [3, 4]]\}
    }
    
    \textbf{Number of vertices (n):}
    \textit{ 
    \{`n': 5\}, \{`n': 6\}, \{`n': 4\}, \{`n': 4\}, \{`n': 6\}
    }
    
    \textbf{Output:}
    \textit{  
    \{`label': 3\}, \{`label': 3\}, \{`label': 3\}, \{`label': -1\}, \{`label': 3\}
    }
    
\end{tcolorbox}

\begin{lstlisting}[caption={Code 1: Code output generated by the SSD-enhanced GPT-4o.}, label={list:ssd}]
class Solution:
  def findShortestCycle(self, n:int, edges:List[List[int]]) -> int:
    adjacency_list = [[] for _ in range(n)]
    for u, v in edges:
    ......
    def bfs(start):
        ......
        while queue:
            node, parent, depth = queue.popleft()
            for neighbor in adjacency_list[node]:
                if neighbor == parent:
                    continue 
                if neighbor in visited:
                    # Cycle detected
                    cycle_length = depth + visited[neighbor] + 1
                    if cycle_length >= 3:
                        shortest_cycle = min(shortest_cycle, cycle_length)
                else:
                    visited[neighbor] = depth + 1
                    queue.append((neighbor, node, depth + 1))
    for i in range(n):
        bfs(i)
    ......
TestCase: n=6, edge=[[4,2],[5,1],[5,0],[0,3], \
        [5,2],[1,4],[1,3],[3,4]]
Output: 3 (Correct)
\end{lstlisting}

\begin{lstlisting}[caption={Code 2: Code output generated by the CoT-based GPT-4o (hallucinated).}, label={list:cot}]
class Solution:
    def findShortestCycle(self, n:int, edges:List[List[int]]) -> int:
        graph = defaultdict(list)
        ......
        def bfs(start):
        dist = [-1] * n
        dist[start] = 0
        queue = deque([(start, -1)]) 
        while queue:  
            current, parent = queue.popleft()
            # Hallucination Happens!
            for neighbor in graph[current]:
                if dist[neighbor] == -1:
                    dist[neighbor] = dist[current] + 1
                    queue.append((neighbor, current))
                elif neighbor != parent: 
                    return dist[current] + dist[neighbor] + 1
        return float('inf')
    shortest_cycle = float('inf')
    for i in range(n):
        shortest_cycle = min(shortest_cycle, bfs(i))
    ......
TestCase: n=6, edge=[[4,2],[5,1],[5,0],[0,3], \
        [5,2],[1,4],[1,3],[3,4]]
Output: 4 (Incorrect)
\end{lstlisting}
\vspace{0.2pt}

We use GPT-4o as the demo model. The generated code details are shown in Code ~\ref{list:ssd} and ~\ref{list:cot}. More reasoning details and examples can be found in the Appendix~\ref{section:completed_reasoning_example}, ~\ref{sec:appendix-graph-concept} and ~\ref{sec:appendix-graph-concept-solving}. We observe that the \methodname{}-enhanced method GPT-4o reduces hallucinations and improves graph problem-solving capabilities. In the following, we further evaluate \methodname{} on various graph problems selected from \datasetname{}. The results are shown in Table~\ref{tab:ssd_on_concepts_and_3_new_graphs}, which compares the performance of GPT-4, GPT-4o, Gemini-pro, and Claude-3-sonnet. More results can be found in Appendix~\ref{sec:appendix-ssd-results}.

\paragraph{Results on Graph Problem Concepts} 
\methodname{} leads to significant improvements across most problem types. For Path Finding, GPT-4 improves from 80.25\% to 88.83\%, GPT-4o from 82.15\% to 88.76\%, Claude-3-sonnet from 85.24\% to 95.56\%, while Gemini-pro shows a slight improvement from 0\% to 0.37\%. 
In Path Planning, GPT-4 and GPT-4o maintain 100\% with or without \methodname{}, while Gemini-pro and Claude-3-sonnet have a notable improvement from 36.78\% to 66.67\%, and 68.96\% to 100\%, respectively.
For Graph Construction, GPT-4 improves from 16.13\% to 19.29\%, GPT-4o from 22.32\% to 25.49\%, Gemini-pro from 12.97\% to 19.04\%, and Claude-3-sonnet from 31.44\% to 36.75\%. In DAG-Related Calculations, GPT-4 and GPT-4o achieve 100\% with \methodname{}, compared to 85.50\% and 90.35\%, respectively, Claude-3-sonnet improves from 54.88\% to 84.15\%, while Gemini-pro from 0\% to 2.44\%. 
The results show that \methodname{} improves performance across all models, highlighting our effectiveness in enhancing LLMs with varying capacities.
GPT-4 and GPT-4o, already strong in tasks, reach near-perfect performance, while Claude-3-sonnet benefits greatly in more complex problems. Gemini-pro shows substantial gains, due to its original lower performance.

\paragraph{Results on Directed and Undirected Graphs}
In directed graphs, GPT-4 improves from 51.25\% to 64.92\%, and GPT-4o sees a slight increase from 91.67\% to 92.31\%. Gemini-pro shows a clear jump from 15.54\% to 40.71\%, and Claude-3-sonnet improves from 66.67\% to 75.00\%.
In undirected graphs, the models generally perform better, with GPT-4 improving from 89.87\% to 92.00\% and GPT-4o from 84.35\% to 92.00\%. Gemini-pro shows a notable improvement from 25.61\% to 64.00\%, and Claude-3-sonnet from 52.91\% to 83.20\%. The results demonstrate that \methodname{} effectively enhances model performance on both directed and undirected graphs, particularly for those with lower baseline performance.

\paragraph{Results on Complex Graphs}
\methodname{} leads to significant improvements across all models for complex graph types. For Small-World graphs, GPT-4 and GPT-4o maintain 100\%, while Gemini-pro improves from 0\% to 68.50\%, and Claude-3-sonnet from 50.00\% to 98.50\%. 
In Erdos-Renyi graphs, Gemini-pro jumps from 33.33\% to 79.33\%, and Claude-3-sonnet from 64.67\% to 95.67\%, with GPT-4 and GPT-4o both reaching 100\%. 
Similarly, in Power-Law graphs, Gemini-pro improves from 33.33\% to 80.00\%, and Claude-3-sonnet from 66.67\% to 100\%.
The results show that \methodname{} significantly enhances models like Gemini-pro and Claude-3-sonnet, which initially perform poorly on complex graphs. \methodname{} enables them to handle diverse structures more effectively.

%% file: limitation.tex

\section*{Limitation}
Despite the strengths of our proposed dataset \datasetname{}, there are certain limitations to consider. The dataset includes 40 coding problems and 36,900 graph samples, which, while comprehensive, is smaller in size compared to other LLM evaluation datasets. Expanding the dataset could offer broader coverage and further insights into model performance across a wider array of graph problems. Additionally, while we evaluate ten prominent LLMs (GPT-3.5, GPT-4, GPT-4o, Claude-3 Sonnet, Gemini-Pro, llama3-8b, llama3-70b, Mixtral-8x7b, Qwen2.5-Coder-32B and DeepSeek-V3), we were unable to include all available models due to the fast-evolving nature of LLM development. This rapid progression in the field may mean that newer models could offer improved performance or demonstrate different behavior on the dataset. However, our dataset is designed to be flexible and general enough to accommodate future LLMs, enabling the research community to use it for evaluating a wide range of models, regardless of advancements in LLM development. Future work can also focus on expanding the dataset size and further refining its categories to address evolving needs in LLM evaluation.

\section*{Ethical Considerations}
In conducting this study, we do not foresee any ethical concerns. Our dataset consists solely of synthetic graph coding problems and generated test cases, ensuring no involvement of personal data or human participants. The development of \datasetname{} and our evaluation of LLMs aimed at improving the understanding of graph problem-solving capabilities in large language models. Furthermore, all data used is anonymous, and the research is compliant with ethical guidelines for responsible AI research. No identifiable information, real-world consequences, or human subjects are involved in this work, thereby minimizing ethical risks.

\section*{Acknowledgment}
This material is based upon work supported by the National Science Foundation under grant no. 2229876 and is supported in part by funds provided by the National Science Foundation, by the Department of Homeland Security, and by IBM.
Any opinions, findings, and conclusions or recommendations expressed in this material are those of the author(s) and do not necessarily reflect the views of the National Science Foundation or its federal agency and industry partners.

%% file: code.tex
\section{Complete Reasoning Example}\label{section:completed_reasoning_example}
Here, we provide further details on the \methodname{}-enhanced graph reasoning and CoT-based graph reasoning: Figure~\ref{box:apd-ssd} and Figure~\ref{box:Cot-based_GPT-4o} (``\methodname{}-enhanced GPT-4o's Graph Reasoning'' and ``CoT-based GPT-4o's Graph Reasoning'').

\definecolor{green}{HTML}{B7C47D}

\begin{figure}[h!]

\begin{tcolorbox}[colback=white, colframe=green, title=\ding{52} Complete \methodname{}-enhanced GPT-4o's Graph Reasoning, coltitle=black, fontupper=\fontsize{7pt}{2pt}\selectfont]

\textbf{<Problem Clarification>}

The goal is to find the length of the shortest cycle in a bi-directional graph, where:

1. Each vertex is labeled from 0 to \( n-1 \).\\
2. Edges are given as a list of pairs, each representing a connection between two vertices.

\textbf{Inputs:} \( n \): Number of vertices in the graph. \textit{edges}: List of edges where each edge is a pair \([u_i, v_i]\) indicating a bi-directional connection between vertices \( u_i \) and \( v_i \).

\textbf{Outputs:} The function returns the length of the shortest cycle if it exists, otherwise returns -1.

\textbf{Constraints:} Each pair of vertices has at most one edge between them. No self-loops (edges where a vertex has an edge to itself).

\textbf{<Problem Breakdown>}\\
\textbf{1. Graph Representation:} Convert the list of edges into an adjacency list to efficiently represent the graph.\\
\textbf{2. Cycle Detection:} Implement a strategy to find and determine the length of the shortest cycle using BFS or DFS.

\textbf{Steps to Solve:}\\
1. Create an adjacency list from the given edges.\\
2. Traverse the graph from each vertex using BFS to find the shortest cycle involving that vertex.\\
3. Track and update the shortest cycle found during the traversals.\\

\textbf{<Solution Formulation>}\\
\textbf{Algorithm:}
\begin{itemize}[leftmargin=*]
    \item \textbf{Breadth-First Search (BFS):}
    \begin{itemize}[leftmargin=*]
        \item Initialize an adjacency list from the edges.
        \item For each vertex, perform a BFS to detect the shortest cycle starting from that vertex:
        \begin{itemize}[leftmargin=*]
            \item Keep track of visited nodes and the level of each node during the BFS using a queue.
            \item When a visited node is encountered, calculate the potential cycle's length and update the shortest cycle length if this cycle is shorter than the previously found ones.
        \end{itemize}
    \end{itemize}
    \item \textbf{Handling Edge Cases:}
    \begin{itemize}[leftmargin=*]
        \item If no cycles are found after all vertices are processed, return -1.
    \end{itemize}
\end{itemize}

\textbf{<Program Implementation>}\\
\textbf{(the code is shown in Code (\ref{list:ssd_appendix})).}
\end{tcolorbox}
\caption{Graph Reasoning by GPT-4o with SSD.}\label{box:apd-ssd}
\end{figure}

\begin{figure}[h!]
\begin{tcolorbox}[colback=white, colframe=green, title=\ding{56} CoT-based GPT-4o's Graph Reasoning, coltitle=black, fontupper=\fontsize{7pt}{2pt}\selectfont]
\textbf{[GPT-4o’s Response]}\\
To solve the problem of finding the shortest cycle in an undirected graph, we can use the Breadth-First Search (BFS) technique. BFS is particularly suitable for this task because it explores all vertices at the present depth level before moving on to vertices at the next depth level, allowing us to find the shortest paths efficiently.\\

Here's an approach to implement this solution:\\
\textbf{(the code is in Code (\ref{list:cot_appendix}), where hallucination happens.)}\\
Graph Representation: Represent the graph using an adjacency list.
BFS from Each Node: For each node, perform a BFS to find the shortest cycle starting and ending at that node.
Cycle Detection: During the BFS, if we encounter a previously visited node that is not the parent of the current node, we have detected a cycle.
Track Minimum Cycle Length: Track the length of the shortest cycle found during all BFS traversals.
Edge Cases: If no cycle is found after all BFS traversals, return -1.

\end{tcolorbox}
\caption{Graph Reasoning by GPT-4o with CoT.}\label{box:Cot-based_GPT-4o}
\end{figure}

\newpage
The detailed code for \methodname{}-enhanced Graph Reasoning and CoT-based Graph Reasoning is provided in Listings 1 and 2, respectively. Notably, the instances of hallucinations in the CoT-based Graph Reasoning are highlighted for clarity.

\begin{lstlisting}[caption={Code 3: Code generated by the SSD-enhanced GPT-4o.}, label={list:ssd_appendix}]
class Solution:
    def findShortestCycle(self, n:int, edges:List[List[int]]) -> int:
        adjacency_list = [[] for _ in range(n)]
        for u, v in edges:
            adjacency_list[u].append(v)
            adjacency_list[v].append(u)

        shortest_cycle = float('inf')

        def bfs(start):
            nonlocal shortest_cycle
            queue = deque([(start, -1, 0)])  
            visited = {start: 0} 

            while queue:
                node, parent, depth = queue.popleft()
                for neighbor in adjacency_list[node]:
                    if neighbor == parent:
                        continue 
                    if neighbor in visited:
                        # Cycle detected
                        cycle_length = depth + visited[neighbor] + 1
                        if cycle_length >= 3:
                            shortest_cycle = min(shortest_cycle, cycle_length)
                    else:
                        visited[neighbor] = depth + 1
                        queue.append((neighbor, node, depth + 1))

        for i in range(n):
            bfs(i)

        return -1 if shortest_cycle == float('inf') else shortest_cycle
\end{lstlisting}

\begin{lstlisting}[caption={Code 4: Code generated by the CoT-based GPT-4o (hallucinated).}, label={list:cot_appendix}]
class Solution:
    def findShortestCycle(self, n:int, edges:List[List[int]]) -> int:
        graph = defaultdict(list)
        for u, v in edges:
            graph[u].append(v)
            graph[v].append(u)

        def bfs(start):
            dist = [-1] * n
            dist[start] = 0
            queue = deque([(start, -1)]) 
            while queue:  # Hallucination Happens!
                current, parent = queue.popleft()
                for neighbor in graph[current]:
                    if dist[neighbor] == -1: 
                        dist[neighbor] = dist[current] + 1
                        queue.append((neighbor, current))
                    elif neighbor != parent:  
                        return dist[current] + dist[neighbor] + 1
            return float('inf')
        
        shortest_cycle = float('inf')
        for i in range(n):
            shortest_cycle = min(shortest_cycle, bfs(i))
        
        return -1 if shortest_cycle == float('inf') else shortest_cycle
\end{lstlisting}

\newpage
\subsection{Reasoning Examples of LLM on Directed and Undirected Graphs} For completeness, we show below how the reasoning process proceeds for undirected and directed graphs with GPT-4o.\label{paragraph: LLM_performance_on_directed&undirected_graphs}

\begin{figure}[h!]
    
\begin{tcolorbox}[colback=white, colframe=green, title=\ding{52} GPT-4o Reasoning on \underline{\textbf{Undirected}} Graphs, coltitle=black, fontupper=\fontsize{7pt}{2pt}\selectfont]
\textbf{[Undirected Graph Example]}\\
There is an infrastructure of n cities with some number of roads connecting these cities. Each roads[i] = [ai, bi] indicates that there is a bidirectional road between cities ai and bi. The network rank of two different cities is defined as the total number of directly connected roads to either city. If a road is directly connected to both cities, it is only counted once. The maximal network rank of the infrastructure is the maximum network rank of all pairs of different cities. Given the integer n and the array roads, return the maximal network rank of the entire infrastructure.\\

\textbf{[GPT-4o's Reasoning Steps]}\\
\begin{itemize}[leftmargin=*]
    \item Step 1: Initialize Data Structures\\
        - We need an array degree to store the number of roads connected to each city.\\
        - We also need a set direct\_connection to keep track of whether two cities are directly connected.

    \item Step 2: Populate the Degree Array and Direct Connection Set\\
        - Iterate through each road in the roads array. For each road connecting cities ai and bi:\\
          a. Increment the degree of both ai and bi.\\
          b. Add the pair (ai, bi) (or (bi, ai)) to the direct\_connection set.

    \item Step 3: Calculate the Maximal Network Rank\\
          - Iterate over all pairs of different cities (i, j):\\
            a. Calculate their combined network rank as degree[i] + degree[j].\\
            b. If cities i and j are directly connected, subtract 1 from their combined network rank (because we don’t want to count the direct road twice).\\
          - Keep track of the maximum network rank encountered.

    \item Step 4: Return the Result\\
           - Return the maximum network rank found during the iteration.
\end{itemize}

\end{tcolorbox}
\caption{Reasoning process of GPT-4o on undirected graphs.}
\end{figure}


\begin{figure}[t]
\vspace{-340pt}
\begin{tcolorbox}[colback=white, colframe=green, title=\ding{52} GPT-4o Reasoning on \underline{\textbf{Directed}} Graphs, coltitle=black, fontupper=\fontsize{7pt}{2pt}\selectfont]
\textbf{[Directed Graph Example]}\\
There are a total of numCourses courses you have to take, labeled from 0 to numCourses - 1. You are given an array prerequisites where prerequisites[i] = [ai, bi] indicates that you must take course ai first if you want to take course bi. For example, the pair [0, 1] indicates that you have to take course 0 before you can take course 1. Prerequisites can also be indirect. If course a is a prerequisite of course b, and course b is a prerequisite of course c, then course a is a prerequisite of course c. You are also given an array queries where queries[j] = [uj, vj]. For the jth query, you should answer whether course uj is a prerequisite of course vj or not. Return a boolean array answer, where answer[j] is the answer to the jth query.\\

\textbf{[GPT-4o's Reasoning Steps]}\\
\begin{itemize}[leftmargin=*]
    \item Step 1: Model the Courses as a Graph\\
         - Each course can be represented as a node in a directed graph, and a prerequisite relationship between two courses can be represented as a directed edge.
    \item Step 2: Build the Graph\\
         -  We'll use an adjacency list to represent the graph. The adjacency list for the graph is a dictionary where each key is a course, and its value is a list of courses that depend on it (i.e., courses that can be taken after completing the key course).
    \item Step 3: Precompute Reachability with Floyd-Warshall Algorithm\\
         - Initialize a 2D matrix reachable where reachable[i][j] is True if there's a path from course i to course j.\\
         - Iterate over each prerequisite to update the matrix: if there's a direct prerequisite relation, set the corresponding entry in the matrix to True.\\
         - Use the Floyd-Warshall algorithm to update the matrix:\\
            a. For each intermediate course k, check if reachable[i][k] and reachable[k][j] are both True for any pair of courses (i, j). If so, set reachable[i][j] to True.

    \item Step 4: Return the Results\\
         - For each query [uj, vj], return reachable[uj][vj] as the answer.
\end{itemize}

\end{tcolorbox}
\caption{Reasoning process of GPT-4o on directed graphs.}
\end{figure}